\documentclass[reprint,amsmath,amssymb,aps,prx,longbibliography]{revtex4-2}

\usepackage[latin1]{inputenc}
\usepackage[T1]{fontenc}
\usepackage[english]{babel}
\usepackage{amssymb}
\usepackage{amsmath}
\usepackage{amsfonts}
\usepackage{amstext}
\usepackage{soul}
\usepackage{psfrag}
\usepackage{epsfig}
\usepackage{bbm}
\usepackage{bm}
\usepackage{bbold}
\usepackage{calc}
\usepackage{mathrsfs}
\usepackage{textcomp}
\usepackage{color,hyperref}
\usepackage{mathtools}

\mathtoolsset{showonlyrefs} 
\newcommand{\x}{{\bf x}}
\newcommand{\xel}[2]{x^{(#1)}_{#2}}
\newcommand{\y}{{\bf y}}
\newcommand{\w}{{\bf w}}
\newcommand{\error}{{\bf e}}

\newcommand{\amat}{{\bf A}}
\newcommand{\amatel}{A}
\newcommand{\admat}[1]{{\bf A}^{(#1)}}
\newcommand{\admatel}[1]{{A}^{(#1)}}
\newcommand{\varmat}[1]{{\bf A}^{(#1)}}%{{\bf \tilde A}^{(#1)}}
\newcommand{\varmatel}[1]{A^{(#1)}}
\newcommand{\covmat}{{\bf S}}
\newcommand{\covel}{S}
\newcommand{\ccovmat}[1]{\covmat^{(#1)}}
\newcommand{\ccovel}[1]{\covel^{(#1)}}
\newcommand{\autocor}{{\bf S}}
\newcommand{\nautocor}{S_{ii}}
\newcommand{\cormat}{{\bf R}}
\newcommand{\scormat}{\hat{\cormat}}
\newcommand{\corel}{R}
\newcommand{\sccormat}[1]{\scormat^{(#1)}}
\newcommand{\ccorel}[1]{\corel^{(#1)}}

\newcommand{\noise}{\sigma}
\newcommand{\mnoise}{\eta}
\newcommand{\ctime}{\tau}
\newcommand{\ctimeinv}{\tau^{-1}}
\newcommand{\estctimeinv}{\hat\tau^{-1}}
\newcommand{\maxlag}{\delta}
\newcommand{\estmaxlag}{\hat \delta}

\newcommand{\netsize}{n}
\newcommand{\density}{d_e}
\newcommand{\reciprocity}{r_e}
\newcommand{\source}{v}
\newcommand{\conou}{c}
\newcommand{\con}[1]{c^{(#1)}}

\newcommand{\score}{f}
\newcommand{\namedscore}[1]{\score^{\textrm{(#1)}}}
\newcommand{\estnamedscore}[1]{{\hat \score}^{\textrm{(#1)}}}

\newcommand{\covcov}{\bf M}
\newcommand{\covcovel}{M}

\newcommand{\ouinference}{OUI}
\newcommand{\grangercausality}{GC}
\newcommand{\transferentropy}{TE}
\newcommand{\convergentcrossmapping}{CM}
\newcommand{\laggedcorrelation}{LC}
\newcommand{\confoundingfactor}{LCCF}
\newcommand{\reversecausation}{LCRC}

\newcommand{\numnetworks}{N_\textrm{Net}}
\newcommand{\mia}{\Phi}
\newcommand{\numsamples}{N}
\newcommand{\named}[2]{{#1}^{\textrm{(#2)}}}
\newcommand{\anticlust}{\overline{c}}

\newcommand{\prop}{{\bf K}}
\newcommand{\maxval}[1]{\wedge_{#1}}%{\max\{#1\}+1}%
\newcommand{\dirac}{\delta}
\newcommand\hypergeometric[3]{\hspace{.5mm}_2\hspace{-.25mm}F_1\hspace{-1mm}
\left(\begin{matrix}#1\,\\#2\,\end{matrix}\Bigg\vert\,#3\right)}
\newcommand{\mcf}{\psi}
\newcommand{\ttp}{\zeta}
\newcommand{\TFA}{g}
\newcommand{\TFB}{{\zeta}}
\newcommand{\TFC}{{\phi}}
\newcommand{\lmax}{\hat{\ell}}
\newcommand{\lmin}{\check{\ell}}
\newcommand{\ldiff}{\Delta \ell}

\begin{document}

\title{Network inference via process motifs \\ for lagged correlation in linear stochastic processes}

\author{Alice C.\,Schwarze}
\email[Corresponding author:~]{alice.c.schwarze@dartmouth.edu}
\affiliation{Mathematics Department, Dartmouth College, Hanover, NH, USA}

\author{Sara M.\,Ichinaga}
\affiliation{Department of Applied Mathematics, University of Washington, Seattle, WA, USA}

\author{Bingni W.\,Brunton}
\affiliation{Department of Biology, University of Washington, Seattle, WA, USA}

\begin{abstract}
    A major challenge for causal inference from time-series data is the trade-off between computational feasibility and accuracy. Motivated by process motifs for lagged covariance in an autoregressive model with slow mean-reversion, we propose to infer networks of causal relations via pairwise edge measure (PEMs) that one can easily compute from lagged correlation matrices. Motivated by contributions of process motifs to covariance and lagged variance, we formulate two PEMs that correct for confounding factors and for reverse causation. To demonstrate the performance of our PEMs, we consider network interference from simulations of linear stochastic processes, and we show that our proposed PEMs can infer networks accurately and efficiently. Specifically, for autocorrelated time-series data, our approach achieves accuracies higher than or similar to Granger causality, transfer entropy, and convergent crossmapping---but with much shorter computation time than possible with any of these methods. Our fast and accurate PEMs are easy-to-implement methods for network inference with a clear theoretical underpinning. They provide promising alternatives to current paradigms for the inference of linear models from time-series data, including Granger causality, vector-autoregression, and sparse inverse covariance estimation. 
\end{abstract}

\maketitle

\section{Introduction}

The task of deducing a network's structure from observations of dynamics on the network is a challenging problem with many important applications. For example, many researchers infer gene-interaction networks and/or metabolic networks from data from biological experiments to visualize and understand complex biological systems \cite{Dohlman2019}. The evaluation of clinical trials and marketing tests often involves the inference of causal relationships among a set of observed variables, which one can interpret as a network of causal relationships \cite{Lipkovich2020, Varian2016}. Neuroscientists infer structural and functional connections among neurons or brain regions from neural recordings to inform their understanding of the brain \cite{Hermundstad2013}. Financial networks inferred from recordings of stock and/or currency prices or market transactions are important tools for assessing national and global risks of financial crises \cite{Nicola2020, Baltakys2021}. 

In these and other domains, researchers increasingly use the structural analysis of inferred networks to draw conclusions with wide-reaching impact, including implications for design of medical advances, corporate strategies, and public policy. These researchers come from different scientific backgrounds, and some may lack access to advanced computing resources. We are therefore motivated to develop network inference methods (1) that reconstruct networks from observational data with high accuracy, (2) whose assumptions and limitations are easy to understand, and (3) that are easy to implement and use, even without access to high-performance computing systems.

Researchers from diverse disciplines have proposed many pairwise edge measures (PEMs) for the inference of causal relationships between pairs of variables \cite{Granger1969,Schreiber2000} and various network-inference problems \cite{Baggio2018,Mizuno2006,Steinhauser2008,Friedman2012,Rubido2018,Hermundstad2013,Hlinka2013,Barman2017,Porfiri2018,Novelli2019,Zhang2019,Chan2017,Villaverde2014,Bianco-Martinez2016}. Possibly the simplest PEMs are correlation \cite{Mizuno2006,Steinhauser2008,Friedman2012} and lagged correlation \cite{Rubido2018}. In neuroscience, it is common to refer to networks inferred via correlations or lagged correlations in time-series data as networks of ``functional connectivity'' (as opposed to ``structural connectivity'') to avoid a claim that correlations or lagged correlations indicate causal relationships \cite{Hermundstad2013}. Many PEMs are based on information-theoretic considerations  \cite{Hlinka2013,Barman2017,Porfiri2018,Novelli2019,Zhang2019,Chan2017,Villaverde2014,Bianco-Martinez2016}. These PEMs include mutual information \cite{Hlinka2013, Barman2017}, transfer entropy \cite{Porfiri2018, Novelli2019,Zhang2019}, partial information decomposition \cite{Chan2017}, unique information \cite{James2018}, and variants thereof \cite{Villaverde2014,Bianco-Martinez2016}. A very popular PEM is linear Granger causality, which is a measure of edge likelihood based on fitting a VAR model \cite{Stepaniants2020,McCabe2020}. Other PEMs are based on fitting other coupled dynamical systems, including oscillatory systems \cite{Cecchini2021} and other nonlinear systems \cite{Mangan2016,Lai2017}, to time-series data. A recent survey compared 249 PEMs \cite{Cliff2022}.

The relationship between the network structure and the correlations in a coupled stochastic dynamical system has been the subject of numerous studies across many disciplines \cite{David2001,Friedman2008,Lai2017,Cecchini2021}. Inverse covariance estimation is a popular network-inference strategy that assumes that a time-series data set is a set of observations of a linear stochastic process \cite{David2001}. The method aims to identify a plausible coupling structure via inverting the data's sample covariance matrix. Sparse inverse covariance estimation includes additional steps to ensure that the resulting inferred networks are sparse. Graphical LASSO is an example of a widely-used algorithm for performing sparse inverse covariance estimation \cite{Friedman2008}. For possibly nonlinear systems with linearizable coupling functions, Lai\,(2017) proposed to infer a network's structure via products of lagged covariance matrices and their inverses \cite{Lai2017}. In our paper, we derive process motifs and process-motif contributions for covariance and lagged covariance in a linear stochastic process, and we use these theoretical results to develop new PEMs. Effects of motifs and other aspects of local network structure on inference results has also been the subject of a recent study on network inference from oscillatory dynamics \cite{Cecchini2021}.

We propose to use linear combinations of lagged sample correlation matrices as PEMs to infer the structure of unweighted, directed networks with high accuracy at low computational cost from longitudinal observations of a linear stochastic process when the sampling rate of observations is equal to or larger than the characteristic timescale $\ctime$ of the stochastic process. 
Similar PEMs (e.g., Granger causality and transfer entropy) are associated with much higher computational cost than our proposed approach, and  tend to yield inaccurate results when large sampling rates and/or small characteristic timescales lead to time-series data sets with strong correlations between consecutive observations. In particular, we propose two new PEMs: lagged correlation with a correction for confounding factors (\confoundingfactor{}) and lagged correlation with a correction for reverse causation (\reversecausation{}). One can compute these PEMs very fast and easily; an implementation is available as part of our code repository \cite{code}.

Our paper proceeds as follows. In Section \ref{sec:more-intro}, we give a brief introduction to PEM-based network inference. In Section \ref{sec:background}, we compare two models for linear stochastic processes --- the Ornstein--Uhlenbeck process (OUP) and a vector-autoregressive (VAR) model --- and explain their connection to process motifs.  We introduce our stochastic delay-difference (SDD) model and derive its process motifs for covariance and lagged covariance in Section \ref{sec:sdm}. In Section \ref{sec:els}, we use the contributions of these process motifs to develop our proposed PEMs, \confoundingfactor{} and \reversecausation{}. We demonstrate in Section \ref{sec:results} that, for time-series data that we generate using the SDD model, one or both of our PEMs lead to similarly accurate or more accurate inferred networks than several widely-used PEMs including Granger causality, transfer entropy, and convergent crossmapping. We discuss limitations and possible future directions in Section \ref{sec:discussion}.

\section{Network inference via pairwise edge measures}\label{sec:more-intro}

\begin{figure}
\centering
\includegraphics[trim={1.265cm 3.3cm 1cm 3.2cm}, clip,width=0.45\textwidth]{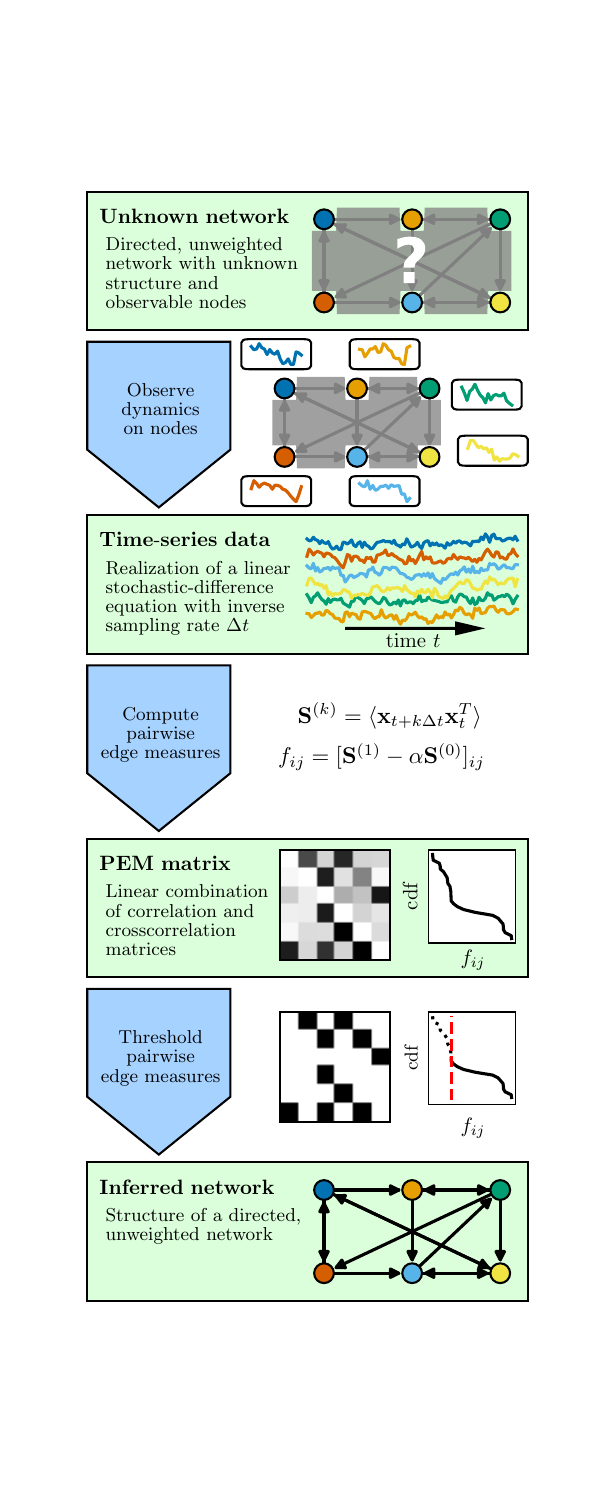}
\caption{{\bf Inferring networks via pairwise edge measures (PEMs).} Our starting point is a directed, unweighted network with unknown structure and observable nodes. Observing node states over time leads to a time-series data set $({\mathbf x}_t)_{t=1,2,3,\dots}$ from which one computes a matrix of PEM values (i.e., ``PEM matrix''). For our proposed inference method, the PEM function is a linear combination of lag-$k$ covariance matrices ${\mathbf S}^{(k)}$ with coefficient $\alpha\in\mathbbm R$. Thresholding the PEM matrix leads to the adjacency matrix of the inferred network.
}
\label{fig:overview}
\end{figure}

In this section, we define the network-inference problem that we consider in this paper. We show a schematic pipeline for PEM-based network inference in Fig.\,\ref{fig:overview}.
Here our starting point is a network with known nodes and unknown structure (i.e., unknown edges), and  we assume that this network is static (as opposed to changing with time), directed, and unweighted. Further, we consider a scenario in which we are able to observe all nodes in the network. 

The first step of network inference via PEMs is to collect observations of all node states over a period of time, yielding a time-series data set. Here we assume that one can model the dynamics that determine the node states by a stable linear stochastic system (e.g., a stable Ornstein--Uhlenbeck process or a stable vector-autoregressive model; see Section \ref{sec:background}) and that the sampling rate $1/\Delta t$ at which one observes the node states is constant and smaller than or equal to the inverse characteristic time $\ctime$ \footnote{The characteristic time of a stable linear system is the time it takes for the norm a perturbation $\Delta \x$ to the system at steady state to decrease by a factor $1/e$.} of the linear stochastic system. 

The next step is to compute PEM values for each node pair from the time-series data. This step yields a ``PEM matrix'', which is a real-valued square matrix with the same number of rows and columns as the adjacency matrix of the network to be inferred. We reason that node pairs with large PEM values are more likely to be connected by an edge than node pairs with low PEM values, so that the PEM values yield a ranking of node pairs by their likelihood of being connected by an edge. 

Either from \textit{a priori} knowledge of the number of edges in a network or via a statistical test, one determines a threshold for the PEM values. The final step is to apply that threshold to the PEM matrix. One interprets the thresholded PEM matrix as the adjacency matrix of the inferred network.

In our paper, we assume that one knows the number of edges in the network \textit{a priori}, and we use that knowledge to determine the threshold for the PEM values. This approach obviates the need to choose one or several statistical tests to determine PEM thresholds, but such tests can be used if prior knowledge is unavailable. It thus allows us to compare the ability of different PEMs to rank node pairs correctly by their edge likelihood regardless of the choice of any statistical test.

\section{Linear stochastic systems and process motifs}\label{sec:background}

In this section, we give brief introductions to a continuous-time and a discrete-time model for linear stochastic dynamics on networks. These models are two limiting cases of the dynamical system that is the theoretical foundation for our proposed PEMs. We also give a brief introduction to process motifs, which we use to derive PEMs (see Section \ref{sec:els}). 

In Subsection \ref{sec:moup}, we briefly review the multivariate OUP, which is a model for continuous-time linear stochastic dynamics on networks. In Subsection \ref{sec:var}, we briefly review VAR models, which model discrete-time linear stochastic dynamics on networks. We compare the OUP with VAR models in Subsection \ref{sec:compare_models}. Finally, in Subsection \ref{sec:pm_intro}, we give a brief introduction to process motifs for linear stochastic processes.

\subsection{The multivariate Ornstein--Uhlenbeck process (OUP)}
\label{sec:moup}
\begin{figure}[!tb]
\centering
\includegraphics[trim={0.11cm 1cm 0.9cm 1.cm}, clip,width=0.45\textwidth]{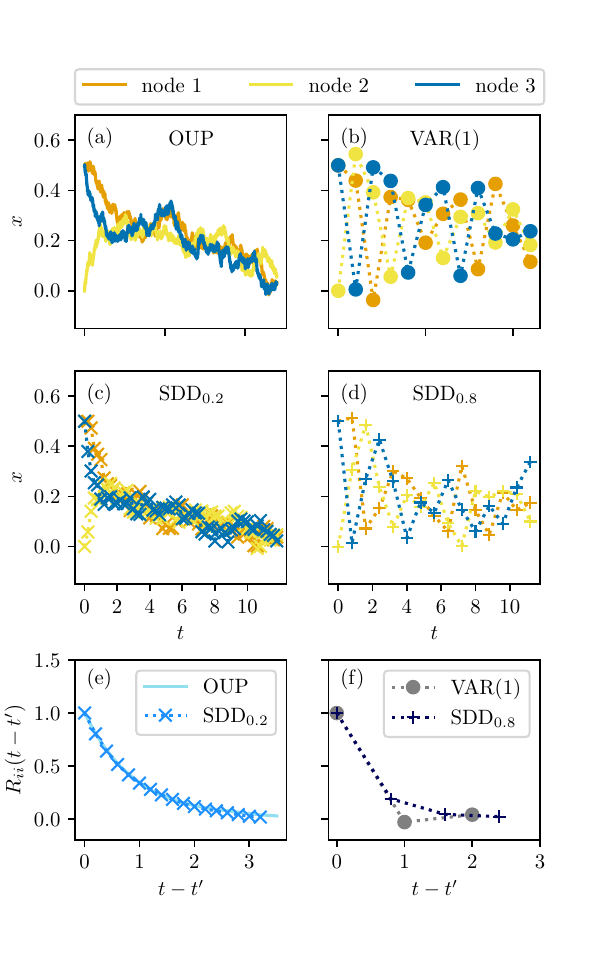}
\caption{{\bf Comparison of continuous-time and discrete-time linear processes on networks.} In (a), we show a realization of an Ornstein--Uhlenbeck process (OUP, see Eq.\,\eqref{eq:moup}) on a unidirectional ring with three nodes. In (b), we show a realization of an order-1 vector-autoregressive model (VAR(1), see Eq.\,\eqref{eq:varp}) on the same network. In (c) and (d), we show realizations of the stochastic delay-difference model (SDD, see Eq.\,\eqref{eq:doup1}) for two different values of the sampling period $\Delta t$. For small $\Delta t$ (panel (c)), the SDD model is similar to the OUP. For large $\Delta t$, the SDD model is similar to the VAR(1) model. We show the respective autocorrelation functions for a single node in panels (e) and (f). For all models, we used coupling strength $\epsilon=0.9$ and noise strength $\noise=0.1$.}
\label{fig:timeseries}
\end{figure}

The OUP is a linear stochastic differential equation that is commonly used to model the transmission of signals on networks. Applications of the OUP include the coupled dynamics of neurons \cite{Barnett2009}, stock prices \cite{Liang2011}, and genes \cite{Rohlfs2013}. In these studies, a neuron, stock, or gene corresponds to a node $i$ in a network and its state (e.g., a neuron's action potential, a stock's price, or a gene's expression level) is described by a variable $\xel{i}{t}$. Collecting the state variables $\xel{1}{t},\xel{2}{t},\xel{3}{t},\dots$ of all nodes in a state vector $\x_t$, the OUP takes the form
\begin{align}
    	d\x_{t}=\ctimeinv(\epsilon\amat-{\bf I})\x_t\, dt+\noise\, d\w_t\,,\label{eq:moup}
\end{align}
where $\amat$ is the adjacency matrix of a network of coupled variables and $\w_t$ is a vector-valued Wiener process that has an $L_2$-norm with an expectation $\langle d\w_t^2\rangle=dt$.
The parameters of the OUP include its characteristic time $\ctime$, its coupling strength $\epsilon$, and its intrinsic noise strength $\noise$.  In Fig.\,\ref{fig:timeseries}\,(a), we show an example of an OUP with $\ctime=1$, $\epsilon=0.9$, and $\noise=0.1$ on a unidirectional ring with three nodes.

The existence of steady-state distributions of the stochastic differential equation in Eq.\,\eqref{eq:moup} depends on the eigenvalues of $\amat$ and the coupling parameter $\epsilon$. If all eigenvalues of $\epsilon\amat-{\bf I}$ are negative, the OUP has a single stable steady state. At that steady state, $\x$ has a multivariate Gaussian distribution with steady-state covariance matrix $\covmat$, which solves the continuous-time Lyapunov equation \cite{Barnett2009, Schwarze2020}
\begin{align}
    0 = (\epsilon\amat-{\bf I})\covmat+\covmat(\epsilon\amat-{\bf I})^T+\frac{\tau\noise^2}{\netsize}{\bf I}\label{moup-coveq}.
\end{align}
One can solve Eq.\,\eqref{moup-coveq} for $\covmat$ and represent its solution as a sum of powers of the adjacency matrix and its transpose \cite{Barnett2009, Schwarze2020},
\begin{align}
	\covmat=\sum_{L=0}^\infty\sum_{\ell=0}^L\conou_{L-\ell,\ell} \amat^{\ell}(\amat^T)^{L-\ell}\,,\label{eq:covmat-ou}
\end{align}
where
\begin{align}
	\conou_{L-\ell,\ell}:=\frac{\ctime\noise^2\epsilon^L}{2^{L+1}\netsize}\binom{L}{\ell}\,.\label{eq:covmat-contributions-ou}
\end{align}

\subsection{Vector-autoregressive (VAR) models}\label{sec:var}
Vector-autoregressive (VAR) models are popular regression models for time-series analysis \cite {Kilian2017}. Their applications include the inference of small networks of causal influences \cite{Hamilton2013, Zhu2020, Goebel2003, Gorrostieta2012, Kook2021, Maciel2020, Jiang2021, Opgen2007, Lu2021}. The inferred networks aim to capture causal relationships between, for example, neurons \cite{Hamilton2013, Zhu2020}, brain regions \cite{Goebel2003, Gorrostieta2012, Kook2021}, genes \cite{Opgen2007, Lu2021}, and the prices of goods or stocks \cite{Maciel2020, Jiang2021}.

An example of a VAR model is the discrete-time stochastic process of the form
\begin{align}
\x_{t} = \sum_{k=1}^{p}\epsilon\varmat{k}\x_{t-k}+\error_t\,,
\label{eq:varp}
\end{align}
where the matrices $\varmat{k}$ describe the influence of the states $\x_{t-k}$ on $\x_t$. To facilitate a direct comparison with the OUP, we include a coupling-strength parameter $\epsilon$ in Eq.\,\eqref{eq:varp}. It is more common to use VAR models to fit time-series data, rather than to simulate synthetic data. When using VAR models to fit time-series data, the error term, $\error_t$, is the residual of the fit. When one can assume that $\error_t$ has a Gaussian distribution, one can use the sample covariance matrix  $\langle\error_t\error_t^T\rangle$ of the stochastic term to measure a VAR model's goodness of fit \cite{Patilea2013}. 

A common way to measure the unconditional linear multivariate Granger causality $\namedscore{\grangercausality{}}_{\y\rightarrow\x}$ from a variable $\y$ to a variable $\x$ is to fit 
(1) the VAR model that regresses over past values of $\x$ and 
(2) the VAR model that regresses over past values of $\x$ and $\y$ 
to observed data. The unconditional linear multivariate Granger causality is given by the difference of log-determinants of  $\langle\error_t\error_t^T\rangle$ for the two models \cite{Stepaniants2020}\,.

When using the model in Eq.\,\eqref{eq:varp}, one assumes that only the $p$ most recent measurements have a non-negligible influence on $\x_t$. The integer $p$ is called the \textit{order} of the VAR model, and it is common to denote a VAR model of order $p$ by VAR$(p)$.
For example, the VAR$(1)$ model from Eq.\,\eqref{eq:var1} can have the form
\begin{align}
    \x_{t} = \epsilon\varmat{1}\x_{t-1}+\error_t\,.\label{eq:var1}
\end{align}
(For the remainder of our paper, we refer to the model in Eq.\,\eqref{eq:var1} as ``the VAR(1) model''.) In Fig.\,\ref{fig:timeseries}\,(b), we show an example realization of the VAR(1) model for which $\varmat{1}$ is the adjacency matrix of a unidirectional ring with 3 nodes. If the error $\error_t$ follows a multivariate Gaussian distribution with covariance matrix 
\begin{align}
\langle\error_t\error_t^T\rangle=\frac{\ctime\noise^2}{\netsize}{\bf I}\label{eq:var_error_cov}
\end{align}
and all eigenvalues $\lambda_i$ of $\varmat{1}$ have negative real parts and absolute values $|\lambda_i|<1$, the VAR$(1)$ model in Eq.\,\eqref{eq:var1} has a stable steady-state distribution that is Gaussian. This stable distribution has an associated steady-state covariance matrix $\ccovmat{0}$ that satisfies the discrete-time Lyapunov equation \cite{Barnett2009}
\begin{align}
    0 = \epsilon\varmat{1}\ccovmat{0}(\epsilon\varmat{1})^T-\ccovmat{0} +\frac{\ctime\noise^2}{\netsize}{\bf I}
\label{eq:var-coveq} 
\end{align}
One can write the solution of Eq.\,\eqref{eq:var-coveq} as a sum of powers of the adjacency matrix and its transpose, 
\begin{align}
    \ccovmat{0} = \frac{\epsilon^2\ctime\noise^2}{\netsize} \sum_{k=0}^\infty\left(\varmat{1}\right)^k\left((\varmat{1})^T\right)^k\,.\label{eq:covmat-var1}
\end{align}
In Eq.\,\eqref{eq:var-coveq} and Eq.\,\eqref{eq:covmat-var1}, we denote the steady-state covariance matrix of a discrete-time process by $\ccovmat{0}$ instead of $\covmat$ (which is our notation for covariance matrices of continuous-time processes). Denoting covariance matrices of a discrete-time processes by $\ccovmat{0}$ makes it possible to denote lag-$k$ covariance matrices of discrete-time processes in Subsection \ref{sec:pm_deriv2} and thereafter by $\ccovmat{k}$.

\subsection{Autocorrelation in the OUP and VAR models}\label{sec:compare_models}
One can view the VAR(1) model as a discrete-time equivalent of an OUP \cite{Ghosh2016}. However, when fitting Eqs.\,\eqref{eq:moup} and \eqref{eq:varp} to data, one can expect that the estimated matrices $\amat$ for the OUP and $\varmat{1}$ models are structurally different. The OUP with $\amat=0$, has an exponentially decaying autocorrelation function,
\begin{align}
    \langle \x_{t'}\x_t^T\rangle=e^{-|t'-t|/\ctime}{\bf I}\,.\label{eq:automoup}
\end{align}
In contrast, the VAR(1) model with $\varmat{1}=0$ has the autocorrelation function, 
\begin{align}
    \langle \x_{t'}\x_t^T\rangle=\dirac_{t,t'}{\bf I}\,,\nonumber
\end{align}
which is $0$ everywhere except at $t=t'$.

The autocorrelation function of the OUP with $\amat=0$ indicates that the OUP is a continuous-time limit of a first-order Markov process. In contrast, the autocorrelation function of the VAR(1) model $\varmat{1}=0$ indicates that the VAR(1) model describes a process in which the value of a state variable is independent of the variable's past values. (Such processes are colloquially called ``zeroth''-order Markov processes \cite{Flack2011, Jost2021}.)
The VAR(1) model describes a first-order Markov process only if $\varmat{1}\neq0$. 

These differences between the OUP and the VAR(1) model influence the interpretation of diagonal elements of $\amat$ and $\varmat{1}$.  A non-zero diagonal element $\varmatel{1}_{ii}$ in $\varmat{1}$ indicates that the state variable $\xel{i}{t}$ is positively or negatively correlated with its most recent past value $\xel{i}{t-1}$. By contrast, a non-zero diagonal element $\amatel_{ii}$ in $\amat$ indicates the state variable $\xel{i}{t}$ depends more strongly or more weakly on its immediate past $\xel{i}{t-dt}$ than one would expect given the exponentially decaying autocorrelation function in Eq.\,\eqref{eq:automoup}. A comparison of panels (a) and (b) in Fig.\,\ref{fig:timeseries} illustrates the qualitative differences between the OUP and the VAR(1) model with $\amat=\varmat{1}$. In panels (e) and (f) of Fig.\,\ref{fig:timeseries}, we use a dark blue curve and a gray curve to indicate the normalized autocorrelation functions $\nautocor(t,t'):=\langle \xel{i}{t}\xel{i}{t'}\rangle/\langle \xel{i}{t}\xel{i}{t}\rangle$ for the OUP with $\amat=0$ and the VAR(1) model with $\varmat{1}=0$. The normalized autocorrelation function for the OUP gradually decays as a function of $t'-t$. The autocorrelation function for the VAR(1) model is only defined where $t'-t$ has integer values. It is equal to 1 for $t-t'=0$ and zero for all other integer values of $t'-t$.

\subsection{Process motifs for linear stochastic processes}\label{sec:pm_intro}
Many researchers have studied motifs in network structure. A \textit{structure motif} $G'$ is a (small) network that (1) occurs as a subnetwork of a network $G\supseteq G'$ and (2) is considered to be important for a function of the network $G$ \cite{Milo2002}. By contrast, a \textit{process motif} is a structured set of walks that (1) occurs on a network $G$ and (2) is considered to be important for a function of $G$ \cite{Schwarze2020}.

\begin{figure}[!t]
\centering
\includegraphics[trim={1.5cm 1.3cm 1.1cm 1.3cm},
clip,width=0.45\textwidth]{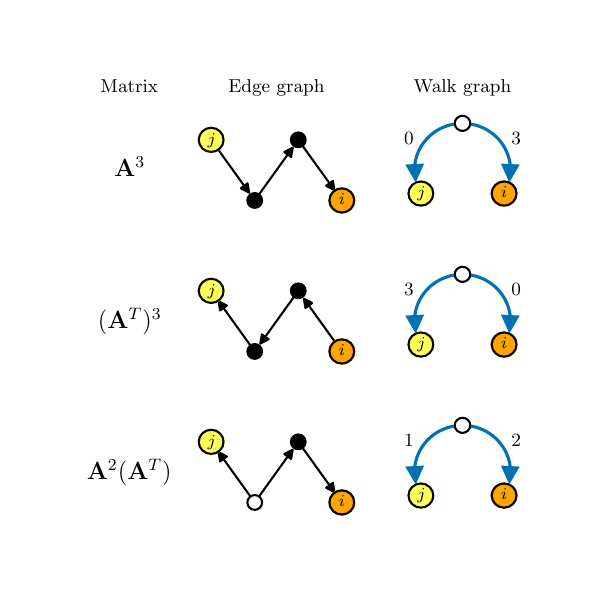}
\caption{{\bf Representations of process motifs.} On the left, we show matrix functions whose elements indicate counts of process motifs for covariance in a network. In the center, we show the examples of small network structures on which the corresponding process motifs can occur. White nodes indicate \textit{source nodes}, which are the starting points of the walks that form a process motif. On the right, we show the corresponding process motifs as walk graphs, in which each curved blue edge with a label ``$k$'' indicates a walk of length $k$ on the network.}
\label{fig:walk_diags}
\end{figure}

For linear dynamical systems, one can express many system properties as a sum of contributions of process motifs \cite{Schwarze2020}. An example are the OUP's steady-state covariances $\covel_{ij}$, for which we show several representations of their process motifs in Fig.\,\ref{fig:walk_diags}. Equation~\eqref{eq:covmat-ou} indicates that one can express the steady-state covariance $\covel_{ij}$ between the nodes $i$ and $j$ as
\begin{align}
    \covel_{ij}=\sum_{L=0}^\infty\sum_{\ell_F=0}^L\conou_{L-\ell_F,\ell_F} \left[\amat^{\ell_F}(\amat^T)^{L-\ell_F}\right]_{ij}\,.
\end{align}
If $\amat$ is an unweighted adjacency matrix, the matrix element $\left[\amat^{\ell_F}(\amat^T)^{L-\ell_F}\right]_{ij}$ corresponds to the number of occurrences of a process motif (i.e., a structured set of walks). The value of $\conou_{L-\ell_F,\ell_F}$ indicates the contribution of each of these process-motif occurrences to $\covel_{ij}$. 

In Fig.\,\ref{fig:walk_diags}, we show the examples of process motifs for steady-state covariance in the OUP. For each example, we show its matrix representation, its representation in a graph of edges, and its representation as a structured set of walks (i.e., walk graphs). If one allows walks to have a 0 length (i.e., one allows the source node $\source$ to coincide with $i$ or $j$), one can consider all process motifs for steady-state covariance in the OUP to be a structured set of two walks that start at the same node $\source$ and end at nodes $i$ and $j$, respectively. A process motif for steady-state covariance is then fully characterized by the lengths of the two walks. We call the walk from $\source$ to $j$ the walk in the ``forward'' direction, and we denote its length by $\ell_F$. We call the walk from $\source$ to $i$ the walk in the ``backward'' direction, and we denote its length by $\ell_B$. The total length of a process motif is $L:=\ell_B+\ell_F$. We use the pair $(\ell_B,\ell_F)$ to denote the process motif whose number of occurrences between $i$ and $j$ is given by $[\amat^{\ell_F}(\amat^T)^{\ell_B}]_{ij}$\,.

\section{A stochastic delay-difference model for a sampled process with slow mean-reversion}\label{sec:sdm}

In this section, we introduce a stochastic delay-difference (SDD) model for sampled time-series data, and we derive its process motifs and process-motif contributions for steady-state covariance and lagged steady-state covariance. In Subsection \ref{sec:model}, we introduce the stochastic difference equation that we use to interpolate between the OUP in Eq.\,\eqref{eq:moup} and the VAR model in Eq.\,\eqref{eq:varp}. In Subsections \ref{sec:pm_deriv} and \ref{sec:pm_deriv2}, we derive and discuss the process motifs for steady-state covariance and lagged steady-state covariance for the stochastic delay-difference equation with Markovian dynamics. In Subsection \ref{sec:extension}, we show how one can extend these results to non-Markovian dynamics. 

\subsection{Stochastic delay-difference equation}\label{sec:model}
Time-series data consists of measurements or observations at a sequence of discrete time points. Whether consecutive observations in the data are correlated depends on the ratio between the characteristic time $\ctime$ of the observed dynamics and the timestep $\Delta t$ between consecutive observations. As a dynamical model for such data, we consider a stochastic delay-difference (SDD) model, 
\begin{align}
    \x_{t} &= \left(1-\Delta t_\ctime\right) \x_{t-\Delta t} + \Delta t_\ctime\sum_{k=1}^p\epsilon\admat{k}\x_{t-k\Delta t}\nonumber\\
    &\phantom{=} +\noise \Delta \w_t\,,\label{eq:doup}
\end{align}
where $\Delta t_\ctime:=\Delta t/\ctime$ is the sampling period in units of $\ctime$ and every element of the vector $\Delta \w_t$ is a $n$-variate 0-mean Gaussian random variable with $\langle\Delta\w_t\rangle^2=\Delta t$. In analogy to the VAR model, we call $p$ the order of the SDD model. When $\Delta t = \ctime = 1$, the SDD model is equivalent to the VAR model in Eq.\,\eqref{eq:varp}.

The SDD model of order $1$ has the form
\begin{align}
    \x_{t} &= \left[{\bf I}+\Delta t_\ctime(\epsilon\admat{1}-{\bf I})\right]\x_{t-\Delta t} +\noise \Delta \w_t\,,\label{eq:doup1}    
\end{align}
which is formally identical to the OUP in Eq.\,\eqref{eq:moup} when one replaces $\Delta t$ with a differential $dt$. We consider the sampling period $\Delta t$ to be a parameter of the SDD model, and it is not infinitesimally small. The SDD model with order $p>1$ allows edges in a network to have associated transmission lags that are multiples of $\Delta t$. In contrast to the OUP, one can thus use the SDD model for non-Markovian dynamics.

In Fig.\,\ref{fig:timeseries}\,(c), we show a realization of the SDD model of order 1 with a small sampling period (i.e., $\Delta t=0.2$) on a unidirectional ring. In panel (d), we show a realization of the same SDD model with a large sampling period (i.e., $\Delta t=0.8$). One can observe that the SDD model with small $\Delta t$ is visually similar to the realization of the OUP shown in Fig.\,\ref{fig:timeseries}\,(a). In Fig.\,\ref{fig:timeseries}\,(e), one can observe that the SDD model with small $\Delta t$ and the OUP have similar autocorrelation functions. The SDD model with large $\Delta t$ is visually similar to a realization of the VAR(1) (see Fig.\,\ref{fig:timeseries}\,(b)) and has a similar autocorrelation function to the VAR(1) (see Fig.\,\ref{fig:timeseries}\,(f)). 

\subsection{Process motifs for covariance}\label{sec:pm_deriv}

In this section, we derive the process motifs and process-motif contributions for steady-state covariance for the SDD model of order 1. The steady-state covariance matrix of the SDD model satisfies the discrete-time Lyapunov equation
\begin{align}
    0 = \prop \ccovmat{0} \prop^T - \ccovmat{0} + \frac{\ctime\noise^2}{\netsize}{\bf I}\,,\label{eq:covmat-prob}
\end{align}
where $\prop:=(1-\Delta t_\ctime) {\bf I} + \Delta t_\ctime\epsilon\admat{1}$. One can express the solution of Eq.\,\eqref{eq:covmat-prob} as 
\begin{align}
    \ccovmat{0} = \sum_{L=0}^\infty\sum_{\ell_F=0}^L\con{0}_{L-\ell_F,\ell_F}(\admat{1})^{\ell_F}((\admat{1})^T)^{L-\ell_F}\,.\:\:\:\label{eq:covmat-sol}
\end{align}
The contributions of process motifs to steady-state covariance in the SDD model are
\begin{align}
    \con{0}_{\ell_B,\ell_F}&= \frac{\ctime\noise^2}{\netsize}\epsilon^{\ell_B+\ell_F}\mcf_{\ell_B,\ell_F}(\Delta t_\ctime)\label{eq:cov_contributions}
\end{align}
where we define 
\begin{align}
    \mcf_{p,q}(z)&:=z^{p+q+1}(1-z)^{|p-q|}\binom{\maxval{p,q}-1}{|p-q|}\nonumber\\
        &\phantom{=}\times\hypergeometric{\maxval{p,q},\,\, \maxval{p,q}}{|p-q|+1}{(1-z)^2}
\end{align}
and use the shorthand
\begin{align}
    \maxval{p,q} :=\max\{p,q\}+1\,.
\end{align}
We derive Eq.\,\eqref{eq:covmat-sol} in Appendix \ref{app:sigma-sol}. The function $\mcf_{p,q}(z)$  captures the effect of nodes ``remembering'' their previous states (i.e., the effect of non-zero correlations between $\xel{i}{t}$ and $\xel{i}{t-\Delta t}$) on the contributions $\con{0}_{\ell_B,\ell_F}$. This effect depends on the autocorrelation function of the process, which depends on the ratio of the sampling period $\Delta t$ and the characteristic time $\ctime$ (see Section \ref{sec:compare_models}). The function $\mcf_{p,q}(z)$ thus depends on $\Delta t_\ctime$, and it is independent of the coupling parameter $\epsilon$ and the noise strength $\noise$. 

As $\Delta t_\ctime$ approaches 0, the steady-state covariance matrix $\ccovmat{0}$ of the SDD model converges to the steady-state covariance matrices of the OUP. We derive this limit in Appendix \ref{app:limit0}. When $\Delta t_\ctime=1$ in Eq.\,\eqref{eq:covmat-sol}, the steady-state covariance matrix of the SDD model is equal to the steady-state covariance matrix of a VAR(1) model with a time-independent error covariance matrix given by Eq.\,\eqref{eq:var_error_cov}. 

One can interpret the elements of the matrix functions $(\admat{1})^\ell((\admat{1})^T)^{L-\ell}$ of the adjacency matrix as counts of process-motif occurrences (see Section \ref{sec:pm_intro}) and the $(i,j)$-th element of the SDD model's steady-state covariance matrix as a sum of these counts --- each weighted by $\con{0}_{\ell_B,\ell_F}$. Comparing Eqs.\,\eqref{eq:covmat-sol} and \eqref{eq:covmat-ou}, one can see that both steady-state covariance matrices are sums of the same matrix functions of the adjacency matrix. The OUP in Eq.\,\eqref{eq:moup} and the SDD model therefore have the same process motifs for covariance, albeit with different contributions --- $c_{\ell_B,\ell_F}$ for the OUP, and $c^{(0)}_{\ell_B,\ell_F}$ for the SDD model. Comparing Eqs.\,\eqref{eq:covmat-sol} and \eqref{eq:covmat-var1}, one can see that the steady-state covariance matrix of the VAR(1) model is a sum of a subset of the matrix functions that contribute to the steady-state covariance matrix of the SDD model. The process motifs for covariance in the VAR(1) model in Eq.\,\eqref{eq:var1} are therefore a subset of the process motifs for covariance in the SDD model.

\begin{figure}[!t]
\centering
\includegraphics[trim={0.cm, 0.6cm, 0.05cm, 1.3cm}, clip,width=0.46\textwidth]{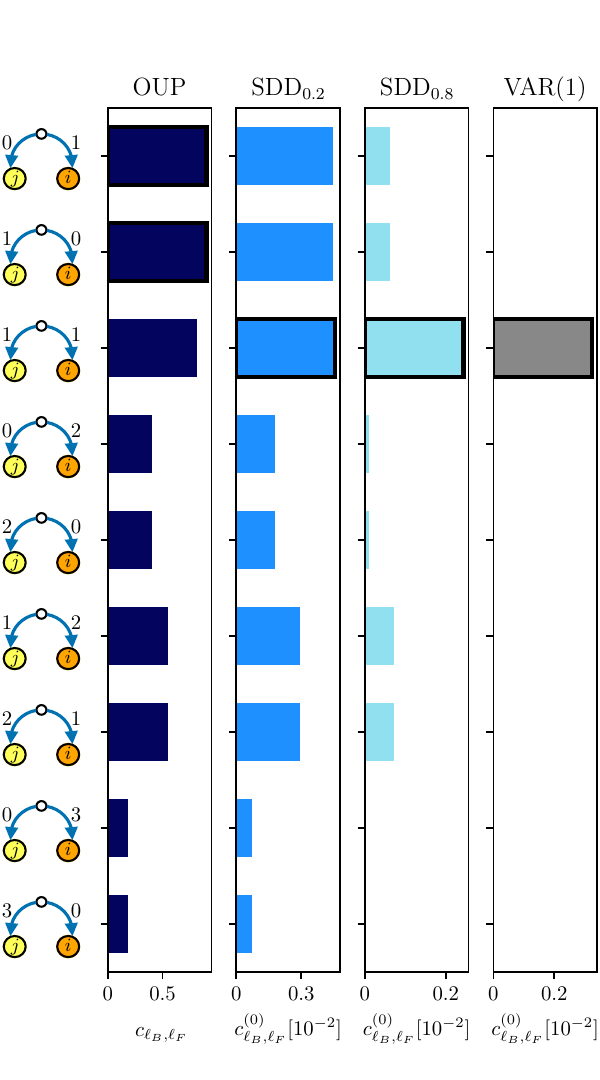}
\caption{{\bf Contributions of process motifs to covariance in linear stochastic processes.} We show the contributions of process motifs to covariance for the Ornstein--Uhlenbeck process (OUP), the stochastic delay-difference model with $\Delta t_\ctime=0.2$ (SDD$_{0.2}$), the stochastic delay-difference model with $\Delta t_\ctime= 0.8$ (SDD$_{0.8}$), and an order-1 vector autoregressive model (VAR(1)). Bars with black outlines indicate the largest contributions to covariance for each stochastic process. We observe that the process motifs (0,1) and (1,0) --- which correspond to true causation and reverse causation \cite{Smith2004, Burgess2021}, respectively ---  contribute most to covariance in the OUP. In the VAR(1) model, the process motif (1,1) --- which corresponds to existence of a confounding factor --- contributes most to covariance. The contributions to covariance in the SDD model interpolate between the contributions to covariance in  the OUP and the VAR(1) model. For all models, we used $\epsilon=0.9$, $\noise=1.0$.}
\label{fig:contributions-cov}
\end{figure}

In Fig.\,\ref{fig:contributions-cov}, we show the contributions of process-motif occurrences to steady-state covariance for the OUP (which corresponds to the SDD model with $\Delta t_\ctime\rightarrow 0$), the SDD model with $\Delta t_\ctime=0.2$, the SDD model with $\Delta t_\ctime= 0.8$, and the VAR(1) model (which corresponds to the SDD model with $\Delta t_\ctime = 1$. One observes that many different process motifs contribute to the steady-state covariances of the OUP. The contributions tend to decrease with increasing difference between the length of the walk to node $i$ and the length of the walk to node $j$ (i.e.,  $|\ell_B-\ell_F|$). The contributions $\con{0}_{\ell_B,\ell_F}$ also tend to decrease with increasing overall length of the process motif (i.e., $\ell_B+\ell_F$). All shown process motifs have larger contributions to the steady-state covariance of the OUP than for the SDD model with a finite sampling rate. As $\Delta t_\ctime$ approaches 1, the SDD model corresponds with the VAR(1) model and the only process motifs with positive contributions to steady-state covariance are process motifs for which the walk to node $i$ and the walk to node $j$ have the same length (i.e.,  $\ell_B=\ell_F$). For the three discrete-time processes that we show in Fig.\,\ref{fig:contributions-cov}, the process motif $(1,1)$ has the largest contribution to steady-state covariance. For the OUP, the process motifs $(0,1)$ and $(1,0)$ have the largest contribution to steady-state covariance. 

To explain why the process motif $(1,1)$ is not the process motif with the largest contribution to steady-state covariance in the OUP, we remark that one can interpret each process motif as a way by which a signal on any node in the network can reach the nodes $i$ and $j$ at the same time. On its way to $i$ and $j$, a signal in the OUP can traverse at most one edge per timestep $\Delta t$. If it does, its amplitude is scaled down by a factor that depends on $\ell_B$ and $\ell_F$ and is proportional to $\epsilon\Delta t_\ctime$. For a process in which node states $\xel{i}{t}$ are correlated with their past $\xel{i}{t'}$ for $t'<t$, signals do not need to traverse an edge in each timestep. One can think of signals ``lingering'' at a node over some number of timesteps. If a signal lingers at a node, its amplitude is scaled down by a factor that depends on $\ell_B$ and $\ell_F$ and is proportional to $(1-\Delta t_\ctime)$. In the process motif $(1,1)$, a signal does not have to linger at any node to contribute to steady-state covariance. Its contribution $\con{0}_{1,1}$ is thus proportional to $(\epsilon\Delta t_\ctime)^2$. In the process motifs $(0,1)$ and $(1,0)$, a signal traverses one edge and lingers for one timestep at one node. Their contributions, $\con{0}_{0,1}$ and $\con{0}_{1,0}$, are thus proportional to $\epsilon\Delta t_\ctime(1-\Delta t_\ctime)$. As $\Delta t$ approaches $0$, traversing an edge leads to a larger decrease on a signal's amplitude than lingering for one timestep at a node. The contribution $\con{0}_{1,1}$ thus approaches 0 faster than the contributions $\con{0}_{0,1}$ and $\con{0}_{1,0}$. One thus has larger contributions of the process motifs $(0,1)$ and $(1,0)$ than the contribution of the motif $(1,1)$ in the OUP.

\subsection{Process motifs for lagged covariance}\label{sec:pm_deriv2}
In this section, we derive the process motifs and process-motif contributions for lagged steady-state covariance in the SDD model.

At steady state, the lagged covariance functions $s_{ij}(t,t'):=\langle\x_{j,t'}\x_{i,t}^T\rangle$ of a stochastic process for given nodes $i$ and $j$ depend only on the difference $\ttp:=t'-t$ between time points. We can thus write the matrix-valued lagged-covariance function 
\begin{align}
    \autocor(\ttp) :=(r(t,t+\ttp))_{ij} = \langle\x_{t+\ttp}\x_{t}^T\rangle\,.
\end{align}
For a discrete-time process, the function $\autocor(\ttp)$ is defined on $\ttp\in\{0,\pm\Delta t, \pm2\Delta t, \dots\}$\,. For any integer $k$, we call the matrix
\begin{align}
    \ccovmat{k} := \autocor(k\Delta t) = \langle\x_{t+k\Delta t}\x_{t}^T\rangle\,.
\end{align}
the \textit{lag-$k$ steady-state covariance matrix} of a discrete stochastic process. 
For positive $k$, the lag-$k$ steady-state covariance matrix of the SDD model of order 1 is connected to the $k'$-lag steady-state covariance matrices with $k'\in\{0,\dots,k-1\}$ via a recursive relationship,
\begin{align} 
    \ccovmat{k} &= \langle \x_{t+k\Delta t}\x_{t}^T\rangle\nonumber\\
    &= \langle [(1-\Delta t_\ctime){\bf I}+\Delta t_\ctime\epsilon\admat{1}]\x_{t+(k-1)\Delta t}\x_{t}^T\rangle\nonumber\\
    &= [(1-\Delta t_\ctime){\bf I}+\Delta t_\ctime\epsilon\admat{1}]\ccovmat{k-1}\,.\label{eq:rec-cross}
\end{align}
From the recursive relationship in Eq.\,\eqref{eq:rec-cross} and the order-1 SDD model's covariance matrix in Eq.\,\eqref{eq:covmat-sol}, it follows that the lag-$k$ steady-state covariance matrix has the form 
\begin{align}
    \ccovmat{k} = \sum_{L=0}^\infty\sum_{\ell_F=0}^L\con{k}_{L-\ell_F,\ell_F}(\admat{1})^{\ell_F}\left((\admat{1})^T\right)^{L-\ell_F}\,,\label{eq:ccovmat1}
\end{align}
where 
\begin{align}
    \con{k}_{\ell_B,\ell_F}:=\con{k-1}_{\ell_B,\ell_F}(1-\Delta t_\ctime) + \con{k-1}_{\ell_B,\ell_F-1}\epsilon\Delta t_\ctime\,.\label{eq:ccov_contributions}
\end{align}
From comparing Eqs.\,\eqref{eq:covmat-sol} and \eqref{eq:ccovmat1}, one can see that the same process motifs contribute to all lag-$k$ steady-state covariance matrices for, $k=0,1,2,\dots$, of the SDD model. Their contributions $\con{k}_{\ell_B,\ell_F}$ to the lag-$k$ steady-state covariance matrix are different for different $k$. The differences between them depend on the parameters $\epsilon$ and $\ctime$ of the SDD model, as well as its sampling period $\Delta t$.

\begin{figure}[!t]
\centering
\includegraphics[trim={0.cm, 0.65cm, 0.05cm, 0.3cm}, clip,width=0.46\textwidth]{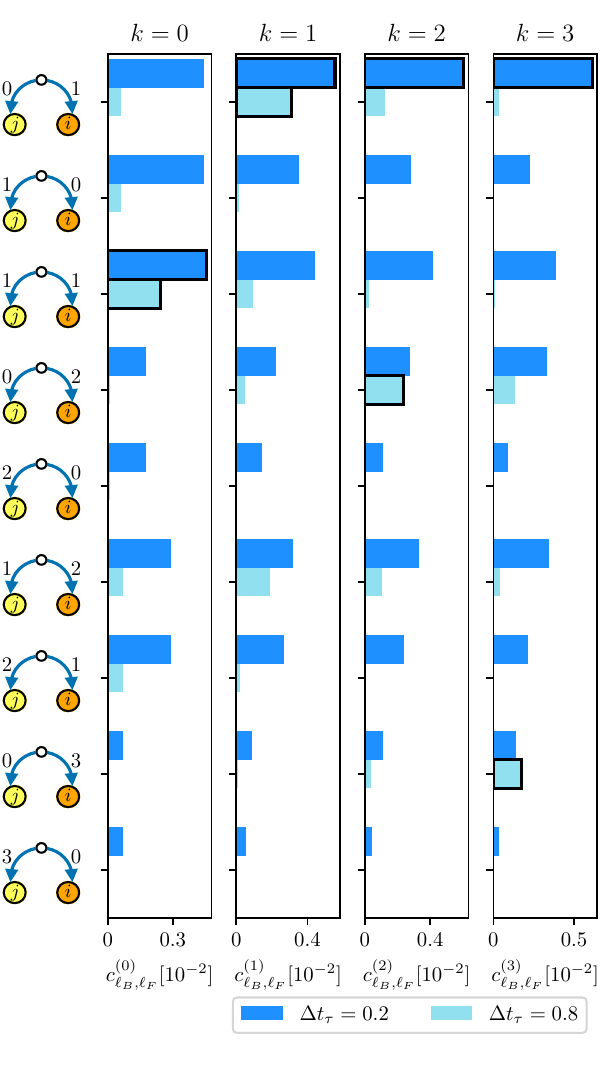}
\caption{{\bf Contributions of process motifs to lag-$k$ covariance.} We show results for the order-1 stochastic delay-difference (SDD) model with $\Delta t_\ctime=0.2$ and $\Delta t_\ctime=0.8$. Bars with black outlines indicate the largest contributions to lag-$k$ steady-state covariance for each stochastic process. For small $\Delta t_\ctime$, the process motif (0,1) tends to contribute the most to lagged steady-state covariance. For large $\Delta t_\ctime$, the process motifs $(0,k)$ tend to contribute the most to the lag-$k$ steady-state covariance. Results are for the SDD model with $\epsilon=0.9$, $\noise=1.0$.}
\label{fig:contributions-cross}
\end{figure}

In Fig.\,\ref{fig:contributions-cross}, we show the contributions of process motifs to the lag-$k$ steady-state covariance matrices with $k\in\{0,1,2,3\}$. We show results for an order-1 SDD model with $\Delta t_\ctime=0.2$ and for an order-1 SDD model with $\Delta t_\ctime=0.8$. One observes that for $\Delta t_\ctime=0.8$ and $k\geq 1$, the largest contributions to elements of the lag-$k$ steady-state covariance matrix are associated with $k$-length walks from node $j$ to node $i$. For $\Delta t_\ctime=0.2$ and $k\geq 1$, the largest contributions are associated with a length-1 walk from $j$ to $i$.

A comparison of the contributions of process motifs to the lag-1 steady-state covariance matrix (see second column of Fig.\,\ref{fig:contributions-cross}) for different $\Delta t_\ctime$ suggests that using the elements of the lag-1 sample covariance matrix, $\ccovmat{1}$, as a PEM tends to yield a higher accuracy for network inference when node states at subsequent timesteps, $\xel{i}{t}$ and $\xel{i}{t+\Delta t}$, are weakly correlated (i.e., $\Delta t_\ctime$ is large) than when node states at subsequent timesteps are strongly correlated (i.e., $\Delta t_\ctime$ is small). When $\Delta t_\ctime$ is large, the contribution of the process motif $(0,1)$ is much larger than the contributions of other process motifs. It follows that a large lagged covariance from node $j$ to node $i$ either suggests a direct connection from $j$ to $i$ or many indirect connections corresponding to other process motifs to covariance. By contrast, for the SDD model with small $\Delta t_\ctime$, the process motifs $(0,1)$, $(1,0)$, and $(1,1)$ have similar contributions to the lagged covariance from node $j$ to node $i$. A large lagged covariance from node $j$ to node $i$ can thus result from many network structures, including an edge from $j$ to $i$, or edges from a third node to both $i$ and $j$.

\subsection{Process motifs for stochastic delay-difference models with order $p>1$}\label{sec:extension}
For the SDD model with order $p>1$, the SDD equation \eqref{eq:doup} includes matrices $\admat{k}$ with $k>1$. The non-zero elements of these matrices indicate edges with lagged signal transmission (i.e., it takes a signal more than one timestep to traverse the edge). We define the \textit{transmission delay} associated with an edge $e$ to be $\maxlag_e=k_e-1$, where $k_e$ is the number of timesteps that it takes for a signal to traverse $e$. 

There are many different reasons why one can expect to observe lagged signal transmission in time-series data. For example, networks that are embedded in a physical space (e.g., road networks, transport networks, vascular networks, natural neural networks, and so on) can include edges with different lengths. When objects or signals travel at a constant speed, they traverse some edges faster than others in these networks. Unobserved nodes are another possible cause for observing delayed signal transmission. For example, when a network includes a path $(j,u,i)$ through an unobserved node $u$, signals traversing this path can contribute positively to the lag-2 correlation from node $j$ to node $i$.

\begin{figure}[!t]
\centering
\includegraphics[trim={0cm 0cm 0cm 0cm}, clip,width=0.46\textwidth]{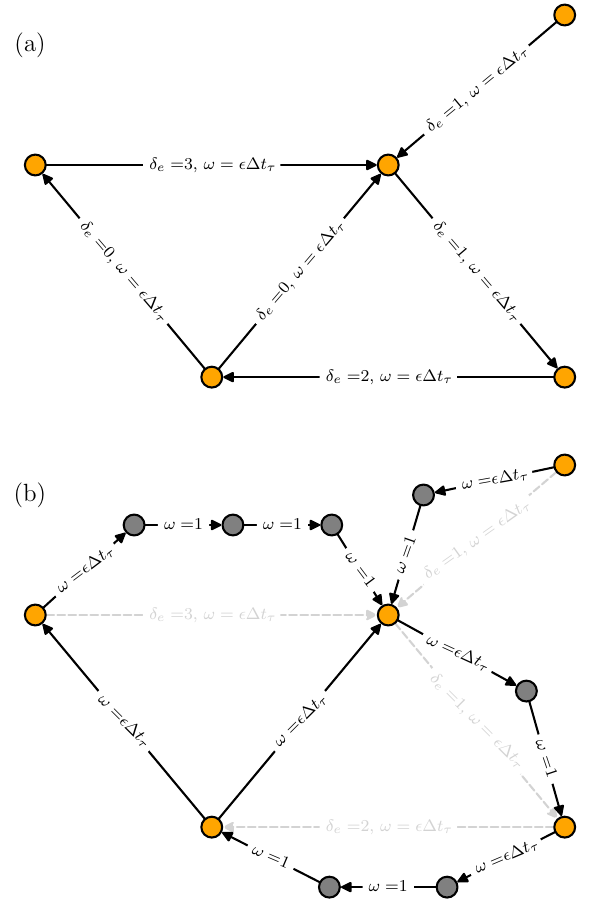}
\caption{{\bf Representation of a stochastic delay-difference (SDD) model of order $k$ as the SDD model of order 1}. In (a), we show a network $G$ of nodes and edges. Edge labels indicate the transmission delays $\maxlag_e$ and weights associated with the SDD model on the network. In (b), we show a corresponding network $G'$, in which we replaced each edge $e$ of $G$ by a path of length $\maxlag_e+1$. Gray nodes are nodes that are in $G'$ but not in $G$. We consider them to be ``unobservable nodes''. We choose edge weights in $G'$ such that the weight of each edge in $G$ is equal to the product of edge weights along the corresponding path in $G'$.}
\label{fig:orderk}
\end{figure}

One can model delayed signal transmission in the SDD model as unobserved nodes in a weighted network. Consider a network $G$ in which each edge $(j,i)$ has a transmission delay $\maxlag_{i,j}$. One can construct another network $G'$ with unobservable nodes such that each edge $(j,i)$ in $G$ corresponds to a path $(j,u_1,u_2,\dots,u_{\maxlag_{ij}}, i)$ of length $\maxlag_{ij}+1$ from $j$ via $\maxlag_{ij}$ unobservable nodes to $i$. To model the decay of a signal's amplitude along edges, one can associate each edge in $G$ with a weight $\epsilon\Delta t_\ctime$. In $G'$, each path that corresponds to an edge in $G$ includes one edge with weight $\epsilon\Delta t_\ctime$. All other edges on the path have a weight of $1$. This choice of edge weights for $G'$ ensures that the amplitude of a signal that traverses an edge in $G$ always decreases by a factor $\epsilon\Delta t_\ctime$, regardless of the transmission delay associated with that edge. In Fig.\ref{fig:orderk}, we give an example of a network $G$ and its associated network $G'$ with unobserved nodes. 

When one constructs $G'$ as described above, the lag-$k$ steady-state covariance matrix of the observable nodes in $G'$ is equal to the lag-$k$ steady-state covariance matrix of $G$ for all integers $k$. A process motif $(\ell_B,\ell_F)$ that has transmission delays on edges has a contribution 
\begin{align}
    \frac{\con{k}_{\ell_B+\bar\maxlag_B,\ell_F+\bar\maxlag_F}}{(\epsilon\Delta t_\ctime)^{\bar\maxlag_B+\bar\maxlag_F}}\,,\label{eq:delaycon}
\end{align}
to the lag-$k$ steady-state covariance in the SDD model of order $p>1$. In Eq.\,\eqref{eq:delaycon}, we use $\bar\maxlag_B$ to denote the sum transmission delays of the edges traversed in the ``backward'' direction (i.e., from the source node $\source$ to node $i$) and $\bar\maxlag_F$ to denote the sum transmission delays of the edges traversed in the ``forward'' direction (i.e., from $\source$ to $j$).

The data shown in Fig.\,\ref{fig:contributions-cross} illustrates that, for large $\Delta t_\ctime$, the process motif $(0,k)$ has the largest contribution to the elements of the 
lag-$k$ steady-state covariance matrix. This observation motivates us to use lag-$k$ correlation matrices to infer edges with transmission delays $k-1$ in the same way that we use lag-1 correlation matrices to infer edges without transmission delay.

\section{From process motifs to pairwise edge measures}
\label{sec:els}

Previously, we discussed that the process-motif contributions to lagged steady-state covariances in the SDD model can suggest whether lag-$k$ steady-state covariances are likely to be good PEMs (see Section \ref{sec:extension}). In the current section, we use process-motif contributions to derive new PEMs that are linear combinations of lag-$k$ steady-state covariances. In Subsection \ref{sec:optimal-els}, we discuss the combination of process-motif contributions that would be an ideal PEM for network inference from the SDD model. In Subsection \ref{sec:els-proposals1}, we propose two PEMs for network inference from the SDD model; one PEM that corrects lag-1 steady-state covariance for confounding factors and one PEM that corrects lag-1 steady-state covariance for reverse causation. In Subsection \ref{sec:estimators}, we explain how we compute these PEMs from time-series data. 

\subsection{An ideal edge-likelihood score function}\label{sec:optimal-els}
An ideal PEM would indicate edges with certainty. A hypothetical example of such an ideal score would be $\namedscore{ideal}_{ij}$ with
\begin{align}
    \namedscore{ideal}_{ij} = \left\{\begin{matrix}1 & \textrm{if $\admatel{1}_{ij}>0$}\\0 & \textrm{otherwise\,.}\end{matrix}\right.\label{eq:ideal}
\end{align} 
Using $\namedscore{ideal}_{ij}$ as a PEM, one would be able to infer networks from observations of the SDD model of order 1 with an accuracy of $1$.

\begin{figure}[!t]
\centering
\includegraphics[trim={0.0cm, 0.6cm, 0.005cm, 1.1cm}, clip,width=0.46\textwidth]{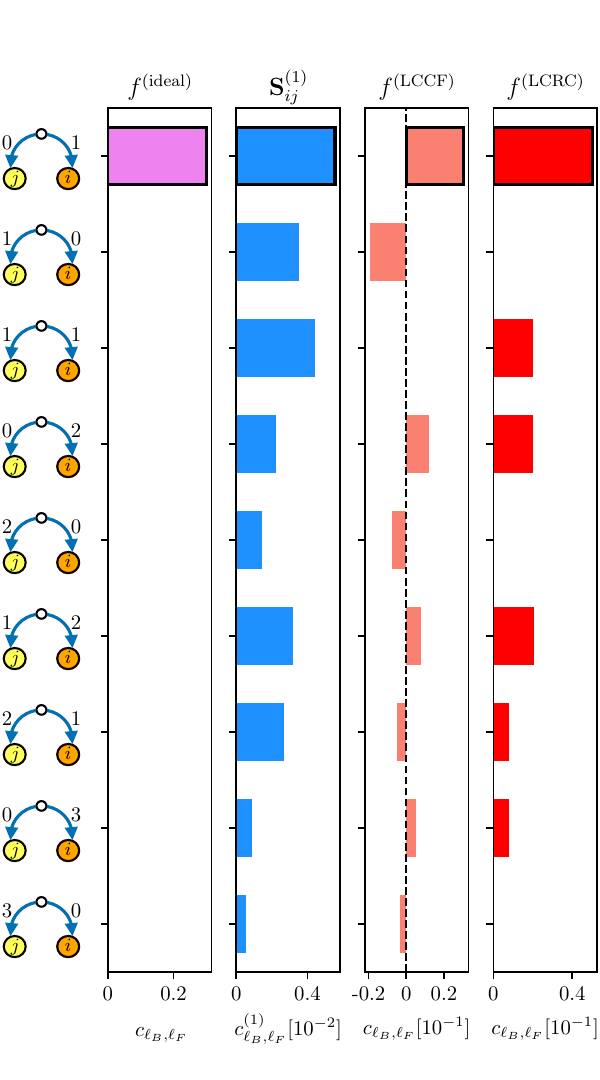}
\caption{{\bf Contributions of process motifs to pairwise edge measures (PEMs).} We show contributions of process motifs for a hypothetical ideal PEM $\namedscore{ideal}_{ij}$, the lag-1 steady-state covariance matrix, lag-1 steady-state covariance with a correction for confounding factors, $\namedscore{\confoundingfactor}_{ij}$, and lag-1 steady-state covariance with a correction for reverse causation, $\namedscore{\reversecausation}_{ij}$. The process motif $(1,1)$ does not contribute to $\namedscore{\confoundingfactor}_{ij}$. The process motifs $(k,0)$ for $k=1,2,3$ do not contribute to $\namedscore{\reversecausation}_{ij}$. Results shown are for the stochastic delay-difference (SDD) model of order-1 with parameters $\Delta t_\ctime=0.5$, $\epsilon=0.9$, and $\noise=1.0$.}
\label{fig:contributions-els}
\end{figure}

In the left column of  Fig.\,\ref{fig:contributions-els}, we show the contributions of process motifs to the ideal PEM $\namedscore{ideal}_{ij}$. The only process motif that has a nonzero contribution to $\namedscore{ideal}_{ij}$ is the process motif $(0,1)$. A comparison of the process-motif contributions for $\namedscore{ideal}_{ij}$ to the process-motif contribution for $\ccovmat{1}$ (see Fig.\,\ref{fig:contributions-cross}) illustrates why lag-$k$ covariance is not an ideal PEM. Many process motifs contribute positively to lag-$k$ covariance. The process motif $(0,1)$ has the largest contribution to lag-1 covariance, but the motifs $(1,0)$ and $(1,1)$ can have almost equally large contributions for small $\Delta t_\ctime$. 

One can think of process motifs for covariance as defining a basis of an infinite-dimensional ``process-motif contribution space'' and use the contributions $\con{k}_{\ell_B,\ell_F}$ of process motifs to a PEM to construct a ``contribution vector'' for the PEM in that space. We use such contribution vectors to compare PEMs. For example, the contribution vector for the ideal PEM $\namedscore{ideal}_{ij}$ is only supported on the process motif $(0,1)$. For $\Delta t_\ctime<1$, the contribution vectors for lag-$k$ steady-state covariance of the SDD model of order 1 are supported on all process motifs. As $\Delta t_\ctime$ approaches one, the contribution vector for lag-$k$ steady-state covariance localizes on process motifs $(\ell_B, \ell_F)$ with $\ell_F=\ell_B+k$.

We are not aware of any function that one can compute in practice from a time-series data set and that fulfills our criterion for being an ideal PEM (i.e., it has a contribution vector with support only on the process motif $(0,1)$). However, one can combine the contribution vectors of lag-$k$ steady-state covariances to construct PEMs whose contribution vectors  localize more strongly on $(0,1)$ than the contribution vectors of lag-$k$ steady-state covariance for any $k=0,1,2,\dots$. We propose two such constructions in Subsection \ref{sec:els-proposals1}.

\subsection{Correcting lag-1 covariance for confounding factors and reverse causation}\label{sec:els-proposals1}

To construct a PEM that is localized on the process motif $(0,1)$, we use linear combinations of contribution vectors for lag-$k$ steady-state covariances. In principle, one can use any number of lag-$k$ steady-state covariances with any values of $k$ to construct a PEM. However, when using lagged sample covariance matrices to approximate lagged steady-state covariance matrices, the sampling errors of each lagged sample covariance matrix can lead to errors of computed PEM values, especially when $k$ is large.

To minimize these sources of error, we construct PEMs that are linear combinations of few lag-$k$ correlation matrices that have small $k$. Specifically, we consider candidate PEMs of the form
\begin{align}
    \score = \ccovmat{1}-\alpha\ccovmat{0}\label{eq:general-proposal}
\end{align}
for some non-negative $\alpha$.

\subsubsection{Correction for confounding factors}
Most process motifs for covariance and lag-1 covariance in the SDD model include two walks of positive length and thus three distinct nodes; a node $i$, a node $j$, and a source node $\source$ from which signals can reach both $i$ and $j$. The state of $\source$ is a confounding factor for the node pair $(j,i)$ because $\source$ can lead to an increase in covariance and lagged covariance between the states of the nodes $i$ and $j$ even when there is no path from $i$ to $j$ and no path from $j$ to $i$. We thus refer to the process motifs for covariance and lagged covariance with three distinct nodes as \textit{confounder motifs}. To construct a PEM with a contribution vector that is localized on the process motif $(0,1)$, we use the steady-state covariance matrix $\ccovmat{0}$ to correct for the confounder motif $(1,1)$, which is the confounder motif that contributes most to lag-1 steady-state covariances. 
To do so, define the PEM 
\begin{align}
    \namedscore{\confoundingfactor{}} := \ccovmat{1}-\named{\alpha}{\confoundingfactor{}}\ccovmat{0}\,,
    \label{eq:lcf0}
\end{align}
with a correction term \footnote{We demonstrate the simplification steps used in the derivation of the correction factor in our supplementary code repository \cite{code} using Python's symbolic-mathematics library \texttt{sympy}.}
\begin{align}
    \named{\alpha}{\confoundingfactor{}}&:=\con{1}_{1,1}/\con{0}_{1,1}\\
    & = 1-\Delta t_\ctime \left(1-\frac{\mcf_{1,0}(\Delta t_\ctime)}{\mcf_{1,1}(\Delta t_\ctime)}\right)\label{eq:lcf0-c0}\\
    &= \frac{2\,(1-\Delta t_\ctime)}{2-2\Delta t_\ctime+\Delta t_\ctime^2}\,.\label{eq:lcf0-c}
\end{align}
The PEM in Eq.\,\eqref{eq:lcf0} combines lag-1 steady-state covariance with a correction term that cancels the contributions of the confounder motif $(1,1)$ to lag-1 covariance. We call this PEM \textit{lagged correlation corrected for confounding factors} (\confoundingfactor{}), and we show the contributions of several process motifs in the third column of Fig.\,\ref{fig:contributions-els}. 

\subsubsection{Correction for reverse causation}
Some process motifs for lag-1 steady-state covariance from $j$ to $i$ include only two distinct nodes and a single walk from $i$ to $j$. This means that signal transmission from $i$ to $j$ can lead to an increase of the lag-1 steady-state covariance from $j$ to $i$ and thus to a false-positive detection of an edge from $j$ to $i$. Some researchers refer to this phenomenon as \textit{reverse causation} \cite{Smith2004, Burgess2021}. When $\Delta t_\ctime$ is small, reverse causation can impact the accuracy of network inference because the difference between the contributions of the process motif $(0,1)$ and $(1,0)$ to lag-1 steady-state covariance is small. This small difference between $\con{1}_{0,1}$ and $\con{1}_{1,0}$ makes it difficult to infer the direction of an edge in a network accurately from the lag-1 steady-state covariance matrix of the SDD model with small $\Delta t_\ctime$. A large lag-1 steady-state covariance from a node $j$ to a node $i$ can be caused by an edge $(j,i)$, but for small $\Delta t_\ctime$, it is almost equally likely that it is caused by an edge in the reverse direction (i.e., an edge $(i,j)$). 
To mitigate false-positive edge detections due to reverse causation, we define the PEM
\begin{align}
    \namedscore{\reversecausation{}} := \ccovmat{1}-\named{\alpha}{\reversecausation{}}\ccovmat{0}\,,
    \label{eq:lrc0}
\end{align}
with a correction term 
\begin{align}
    \named{\alpha}{\reversecausation{}}&:=\con{1}_{1,0}/\con{0}_{1,0}\\
        & = 1-\Delta t_\ctime\,.\label{eq:lrc0-c}
\end{align}
The correction term $\named{\alpha}{\reversecausation{}}$ cancels the contributions of reverse causation to lag-1 steady-state covariance. We call the PEM in Eq.\,\eqref{eq:lrc0} \textit{lagged correlation corrected for reverse causation} (\reversecausation{}), and we show the contributions of several process motifs in the fourth column of Fig.\,\ref{fig:contributions-els}. 

\subsection{Computing pairwise edge measures from time-series data}\label{sec:estimators}
In the OUP, the SDD process, and the VAR model, the process motifs that contribute to covariance are a subset of the process motifs that contribute to correlations
\begin{align}
    \ccorel{k}_{ij}:=\frac{\ccovel{k}_{ij}}{\ccovel{k}_{ii}\ccovel{k}_{jj}}\,,
\end{align}
for $k=0,1,2,\dots$. In the OUP, the process motifs for covariance tend to be the process motifs that contribute the most to correlation \cite{Schwarze2020}. We thus expect that the correction factors that are needed to correct lag-1 steady-state correlation matrices for confounding factors or reverse causation are very similar to the correction factors that we proposed in Eqs.\,\eqref{eq:lcf0-c} and \eqref{eq:lrc0-c}. Using sample covariance matrices as estimators of steady-state covariance matrices and sample correlation matrices as estimators of steady-state correlation matrices for synthetic time-series data (see Section \ref{sec:results}), we found in initial tests that we can infer networks with higher accuracy using correlations than when using covariances. We thus use the following estimators for $\namedscore{\confoundingfactor{}}$ and $\namedscore{\reversecausation{}}$ to infer networks from time-series data;
\begin{align}
\estnamedscore{\confoundingfactor{}}  &:= \sccormat{1}-\named{\alpha}{\confoundingfactor{}}\sccormat{0}\,,\:\:\:\label{eq:lcf}
\end{align}
and
\begin{align}
\estnamedscore{\reversecausation{}}  &:= \sccormat{1} - \named{\alpha}{\reversecausation{}{}}\sccormat{0}\,,\label{eq:lrc}
\end{align}
where $\sccormat{k}$ is the $k$-lag sample correlation matrix of a time-series data set.

To compute the PEM estimators in Eqs.\,\eqref{eq:lcf} and \eqref{eq:lrc}, one needs to know the value of $\Delta t_\ctime:=\Delta t/\ctime$. We assume that the sampling period $\Delta t$ of the time-series data is known. In Appendix \ref{sec:theta_est}, We propose a simple procedure for estimating $\ctime^{-1}$ from a time-series data set. In Appendix, \ref{sec:els-proposals2}, we explain how to generalize Eqs.\,\eqref{eq:lcf} and \eqref{eq:lrc} when one uses the SDD model of order $p>1$ as a model for time-series data.

\section{Network inference via correlation-corrected cross-correlation}\label{sec:results}

In this section, we evaluate the accuracy of our proposed PEMs and several commonly used PEMs for network inference from simulated time-series data. We compare the accuracy of inferred networks from different PEMs as well as the computation time associated with different PEMs. For all PEMs that we consider here, the accuracy of inferred networks depends on parameters of the SDD model that we use to generate time-series data and on structural properties of the ground-truth networks. In Subsection \ref{sec:resultsA}, we investigate how the mean accuracy of inferred networks changes as one varies several parameters of the SDD model. These parameters are the coupling strength $\epsilon$, the characteristic time $\ctime$, and the sampling period $\Delta t$. In Subsection \ref{sec:resultsB}, we present results on the accuracy of inferred networks as one varies structural properties of ground-truth networks. We consider variations in the network size $\netsize$, the edge density $\density$, and the edge reciprocity $\reciprocity$. We also compare the mean accuracy of inferred networks when ground-truth networks are realizations of one of several random-graph models. In Subsection \ref{sec:resultsC}, we investigate the effect of system noise $\noise$ and measurement noise $\mnoise$ on the mean accuracy of inferred networks. Since most of the PEMs formally rely on time-series data being stationary, we also study how nonstationarity of time-series data affects the mean accuracy of inferred networks. We show our results of this study in Subsection \ref{sec:resultsD}. In Subsection \ref{sec:resultsE}, we compare the computation time for different PEMs as one varies the size $\netsize$ of the ground-truth network, the number $\numsamples$ of observations, and the maximum edge transmission lag $\estmaxlag$.

 For the comparisons in this section, we consider our two proposed PEMs, as well as five PEMs that researchers have previously used to infer various networks: lag-1 correlation (\laggedcorrelation{}) \cite{Rubido2018}, transfer entropy (\transferentropy{}) \cite{Schreiber2000}, linear Granger causality (\grangercausality{}) \cite{Bressler2011, Granger1969, Wiener1956}, convergent crossmapping (\convergentcrossmapping{}) \cite{Sugihara2012}, and fitting an undirected multivariate Ornstein--Uhlenbeck model (\ouinference) \cite{McCabe2020}. For all of these methods, we use the implementations provided in the Python package \texttt{netrd} \cite{McCabe2020}. 
 
 For generating time-series data, we simulate the SDD model with coupling strength $\epsilon$, characteristic time $\ctime$, sampling period $\Delta t$, and system noise strength $\noise$ on a ground-truth network with $\netsize$ nodes, an edge density $\density$, and an edge reciprocity $\reciprocity$. Unless we state otherwise, we use the same default parameters for all numerical experiments presented in this section. For the network generation, our default network model is a $G(n,m)$ random-graph model and with $\netsize=10$, $\density=0.5$, and $\reciprocity=0.5$. To ensure convergence of simulations and facilitate a comparison of processes with different coupling strengths, we normalize adjacency matrices so that their largest eigenvalue has absolute value $|\lambda_\textrm{max}|=1$. Whenever a sample network from a random-graph model has a nilpotent adjacency matrix, we reject that network and sample a new network from the random-graph model. Our default parameters for simulating the SDD model are $\epsilon=0.9$, $\ctime=1$, $\Delta t=0.5$, and $\noise=0.2$ (see \cite{dpars} for complete list of default simulation parameters).

\subsection{Dependence on parameters of the stochastic delay-difference model}\label{sec:resultsA}

\begin{figure}[!t]
\centering
\includegraphics[trim={0.45cm, 0.0cm, 0.0cm, 0.0cm}, clip,width=0.46\textwidth]{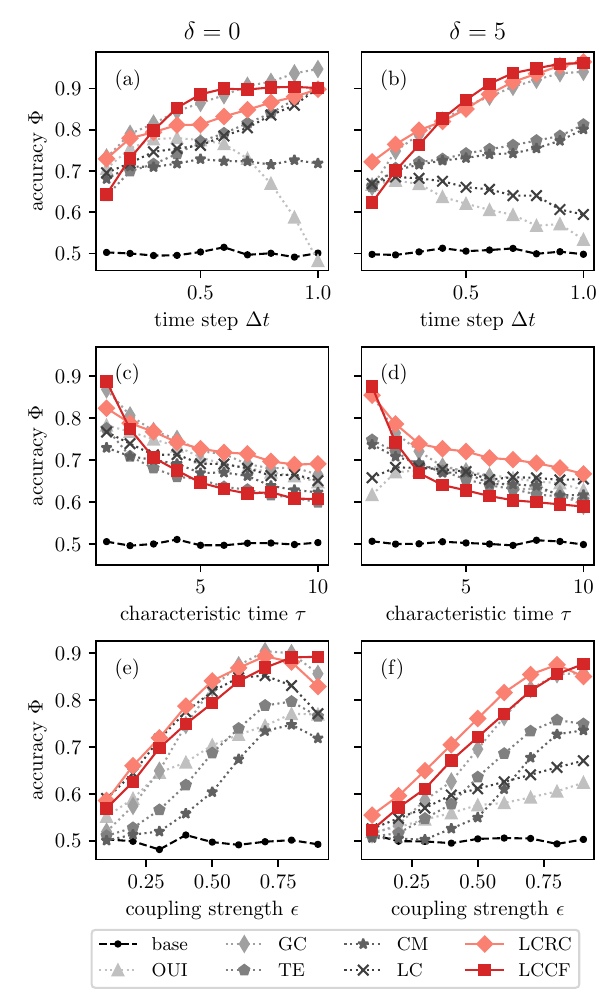}
\caption{{\bf Mean inference accuracy of with varying parameters of the dynamical system.} We show the mean accuracy of inferring networks with Ornstein--Uhlenbeck inference (OUI), linear Granger causality (\grangercausality{}), transfer entropy (\transferentropy{}), convergent cross mapping (\convergentcrossmapping{}), lag-1 correlation (\laggedcorrelation{}), and our proposed methods \reversecausation{} and \confoundingfactor{} (see Eqs.\,\eqref{eq:lrc} and \eqref{eq:lcf}) for inferring networks. The baseline ``base'' corresponds to inferring a network by selecting the correct number of edges uniformly at random. In (a) and (b), we vary the sampling period $\Delta t$. In (c) and (d), we vary the characteristic time $\ctime$. In (e) and (f), we vary the coupling strength $\epsilon$. In (a), (c), and (e), we show the results for dynamics with no transmission lag on edges (i.e., $\maxlag=0$). In (b), (d), and (f), we show results for dynamics with a maximum transmission lag $\maxlag=5$ on edges. We sampled networks from a directed $G(n,m)$ random-graph model. Unless specified otherwise, we use the default parameter values given in \cite{dpars}.}
\label{fig:A1}
\end{figure}

\begin{figure*}[!t]
\centering
\includegraphics[trim={0.0cm, 0.45cm, 0.2cm, 0.365cm}, clip,width=1\textwidth]{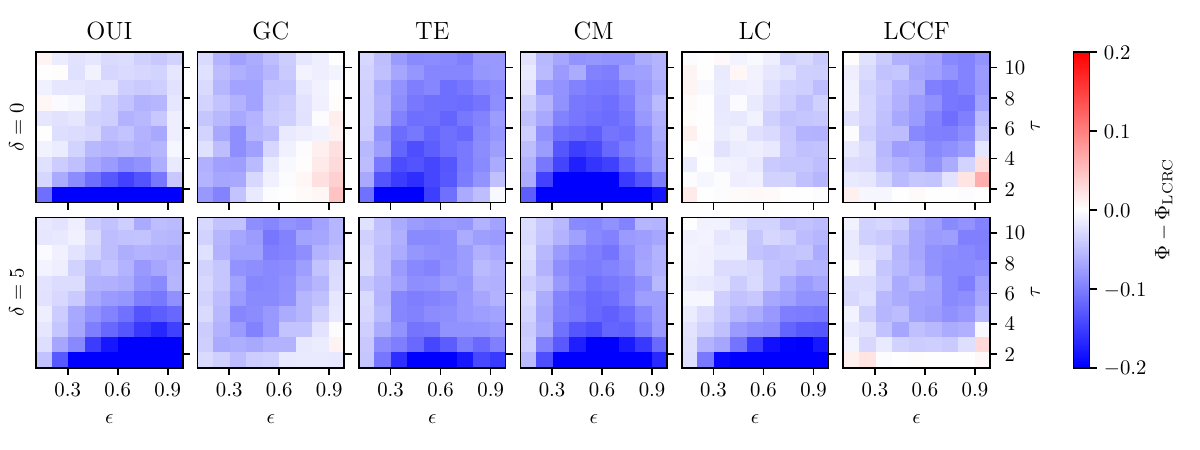}
\caption{{\bf Comparison of inference methods with varying parameters of the dynamical system.} We show the difference between the mean accuracy $\mia_{\textrm{\reversecausation}}$ of inferring networks with \reversecausation{} and the mean accuracy of inferring networks with other PEMs as we vary the coupling strength $\epsilon$, the characteristic time $\ctime$, and the maximum edge transmission lag $\maxlag$. (For values of  other parameters, see \cite{dpars}.) Red pixels indicate that a PEM leads to higher mean accuracy than \reversecausation{}. Blue pixels indicate that a PEM leads to lower mean accuracy than \reversecausation{}. For PEM acronyms, see caption
of Fig.\,\ref{fig:A1}.
}
\label{fig:A2}
\end{figure*}

In Figure \ref{fig:A1}, we show the mean inference accuracy $\Phi$ for inferring realizations of a $G(n,m)$ random-graph model with $\netsize=10$ nodes and $m=45$ (i.e., an edge density $\density=0.5$) as we vary parameters of the SDD model. In Figure \ref{fig:A2}, we show differences in mean inference accuracy between several PEMs and our proposed PEM $\namedscore{\reversecausation}$ as one varies $\epsilon$ and $\ctime$.

\subsubsection{Sampling period}\label{sec:resultsA-dt}

In Figure \ref{fig:A1}\,(a), we show our results for the mean inference accuracy $\Phi$ as one varies the sampling period $\Delta t$ of a SDD model with no transmission lags on edges (i.e., the SDD model of order $1$). We observe that the mean inference accuracy for most of the considered PEMs improves as $\Delta t$ increases. An exception is the PEM for OUI. This PEM relies on the assumption that the time-series data arises from a continuous-time process. This assumption is approximately correct when $\Delta t$ is small (e.g., $\Delta t\lesssim 0.3$ in Fig.\,\ref{fig:A1}\,(a)). When $\Delta t$ is large, the SDD model is very different from to a continuous-time process. Accordingly, assuming that the time-series data is the result of a continuous-time process leads to lower mean inference accuracy for OUI.

When $\Delta t=\ctime$, the correction factors for \confoundingfactor{} (see Eq.\,\eqref{eq:lcf0-c}) and \reversecausation{} (see Eq.\,\eqref{eq:lrc0-c}) have the same value. We thus observe equal mean inference accuracy for \confoundingfactor{} and \reversecausation{} at $\Delta t=1$. The PEM \reversecausation{} leads to a higher mean inference accuracy than \confoundingfactor{} at small $\Delta t$ (i.e., $\Delta t\lesssim 0.2$) and to a lower mean inference accuracy than \confoundingfactor{} at larger $\Delta t$ (i.e., $\Delta t\gtrsim 0.3$). 

Among the PEMs that we compare in Fig.\,\ref{fig:A1}\,(a), at least one of our proposed PEMs leads to the highest or second-highest mean inference accuracy for all considered values of $\Delta t$. For $\Delta t\approx0.3$ and for $\Delta t\gtrsim 0.8$, we obtain the highest mean inference accuracy with \grangercausality{} among our considered network-inference methods. When using linear Granger causality to infer networks from time-series data, one uses a VAR($p$) model for the data. For $\Delta t$ close to 1, the SDD model is similar to a VAR(1) model. The VAR($p$) model that one uses for the calculation of Granger causality is thus a good model for the time-series data, and using \grangercausality{} as a PEM leads to high inference accuracy.

To compare the results for network inference from Markovian dynamics to network inference from non-Markovian dynamics, we also show our results for mean inference accuracy as one varies $\Delta t$ of a SDD model of order 6 (see Fig.\,\ref{fig:A1}\,(b)). For the simulation of the SDD model, we choose edge transmission lags uniformly at random from $\{0,1,\dots,\maxlag\}$, where $\maxlag=5$. 
When moving from the SDD model without transmission lags on edges (i.e., $\maxlag=0$) to the SDD model with $\maxlag=5$, one observes that the mean inference accuracy decreases for \laggedcorrelation{} and \ouinference, which are PEMs that do not account for edge transmission lags.
The mean inference accuracy for \convergentcrossmapping{} is larger for $\maxlag=5$ than for $\maxlag=0$. The other considered PEMs lead to similar or slightly lower mean inference accuracy for $\maxlag=5$ than for $\maxlag=0$. 
This negative effect of edge transmission lags on mean inference accuracy is larger for \grangercausality{} than for \reversecausation{} or \confoundingfactor{}. In Fig.\,\ref{fig:A1}\,(b), either \reversecausation{} or \confoundingfactor{} yields the highest mean inference accuracy among all considered PEMs for all considered values of $\Delta t\in(0,1]$.

\subsubsection{Characteristic time}\label{sec:resultsA-tau}
In Figure \ref{fig:A1}\,(c), we show our results for mean inference accuracy as one varies the characteristic time $\ctime$ of the SDD model with no transmission lags on edges (i.e., a SDD model of order $1$).

We noted in Section \ref{sec:pm_deriv} that the process motifs for covariance and lagged covariance of the SDD model depend on $\Delta t$ and $\ctime$ only through the ratio $\Delta t_\ctime := \Delta t/\ctime$. Accordingly, one observes similar effects when one increases $\ctime$ as when one decreases $\Delta t$. The mean inference accuracy for all considered PEMs decreases with increasing $\ctime$. In Fig.\,\ref{fig:A1}\,(c), the PEMs \reversecausation{}, and \grangercausality{} tend to lead to higher mean inference accuracy than all other PEMs that we considered. For small $\ctime$ (i.e., $\ctime\approx1$), one obtains a higher mean inference accuracy for \confoundingfactor{} than for \reversecausation{} or \grangercausality{}. As one increases $\ctime$, the ranking of the three PEMs changes first such that \grangercausality{} ranks best (for $2\lesssim\ctime\lesssim 3$), then such that \reversecausation{} ranks best (for $\ctime\gtrsim4$).

In Figure \ref{fig:A1}\,(d), we show our results for mean inference accuracy as one varies the $\ctime$ of a SDD model with a maximum transmission lag $\maxlag=5$ on edges. At $\ctime=1$, the PEM \confoundingfactor{} leads to the highest mean inference accuracy among our considered PEMs. However, as one increases the characteristic time to $\ctime=2$, the mean inference accuracy for \confoundingfactor{} drops by almost $15\%$. %, which makes it the least accurate PEM in our comparison for $\ctime\geq3$. 
For $\ctime\gtrsim2$, the PEM \reversecausation{} leads to the highest mean inference accuracy in our comparison of PEMs.

\subsubsection{Coupling strength}\label{sec:resultsA-epsilon}

In Figure \ref{fig:A1}\,(e), we show our results for mean inference accuracy as one varies the coupling strength $\epsilon$ of the SDD model with no transmission lags on edges (i.e., $\maxlag=0$). 

For small to intermediate values of $\epsilon$ (i.e.,  $\epsilon\lesssim0.7$), the mean inference accuracy for all considered PEMs increases monotonically with $\epsilon$. This monotonic increase of $\mia$  with $\epsilon$ is intuitive. At $\epsilon=0$, the SDD model produces only Gaussian white noise, which carries no information on a network's structure. When $\epsilon$ is very small, a signal's amplitude decreases strongly when it traverses an edge. The strong decrease of a signal's amplitude along a single edge makes it difficult to infer edges correctly. As $\epsilon$ increases, signals can traverse edges without a strong decrease of amplitude, which makes it possible to infer networks with an increasingly high accuracy.

For $\epsilon\gtrsim0.7$, the mean inference accuracy for all considered PEMs except \confoundingfactor{} and \ouinference{} decreases with increasing coupling strength. 
A possible explanation for this change in monotonicity with increasing $\epsilon$ is that, as $\epsilon$ approaches 1, there is almost no change in signal amplitude when a signal traverses an edge. Signals thus tend to persist on the network for a very long time, and the assumption that the observed time-series data is stationary becomes inaccurate. Network inference from nonstationary time-series data via PEMs tends to be more challenging than network inference from stationary time-series data and lead to a decreased mean accuracy of inferred networks (see Subsection \ref{sec:resultsE}).

In Fig.\,\ref{fig:A1}\,(e), the PEM $\namedscore{\reversecausation}$ leads to the highest mean accuracy of network inference among all considered PEMs when $\epsilon\lesssim0.6$. For $\epsilon\in[0.7, 0.8]$, Granger causality is the best-performing PEM in our comparison. For $\epsilon=0.9$, \confoundingfactor{} leads to higher mean inference accuracy than any of the other considered methods. 

In Figure \ref{fig:A1}\,(f), we show our results for mean inference accuracy as one varies $\epsilon$ for the SDD model with transmission lags on edges. The results are qualitatively similar to the results for the SDD model without transmission lags on edges. Notably, we observe that both of our proposed PEMs have larger mean inference accuracy than \grangercausality{} at low coupling strengths (i.e., $\epsilon\lesssim0.4$) for networks with and without transmission lags on edges.

\subsubsection{Interplay of characteristic time and coupling strength}

From the results that we presented in Subsections \ref{sec:resultsA-dt}--\ref{sec:resultsA-epsilon}, one can see that the ranking of PEMs based on their associated mean inference accuracy changes with variations in the parameters of the dynamical system that we use to generate time-series data. To give an overview of how changes in the parameters of the SDD model affect the performance of our PEMs, we show heat maps of the difference in mean inference  accuracy between \reversecausation{} and several PEMs in Figure \ref{fig:A2}. Positive values (i.e., red-shaded areas) indicate parameter pairs $(\ctime, \epsilon)$ for which another PEM leads to higher mean inference accuracy than \reversecausation{}. Negative values (i.e., blue-shaded areas) indicate parameter pairs $(\ctime, \epsilon)$ for which \reversecausation{} leads to higher mean inference accuracy than another PEM.

We find that for most combinations of $\epsilon \in (0,1)$ and $\ctime\in[1,10]$, the PEM \reversecausation{} leads to better or comparable mean inference accuracy when compared to  \confoundingfactor{} and all other PEMs that we considered. For dynamics with small to intermediate $\ctime$ (i.e., $\ctime\lesssim5$), large $\epsilon$ (i.e., $\epsilon\gtrsim0.8$) and no transmission lags on edges (i.e., $\maxlag=0$), the mean inference accuracy with \confoundingfactor{} or \grangercausality{} tends to be higher than with \reversecausation{}. For $\maxlag=5$, the mean inference accuracy associated with \reversecausation{} is either higher or less than $1\%$ lower than the mean inference accuracy for \grangercausality{}.

\subsubsection{Effect of lag-parameter mismatch}

\begin{figure}[!t]
\centering
\includegraphics[trim={0.cm, 0.0cm, 0.0cm, 0.65cm}, clip,width=0.46\textwidth]{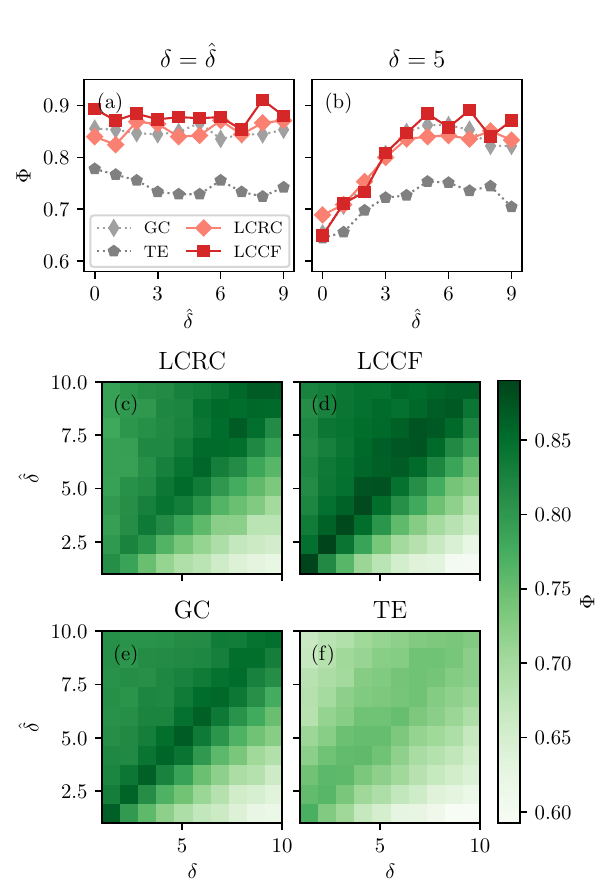}
\caption{{\bf Effect of mismatch between maximum edge transmission lag and expected maximum edge tranmission lag.} In (a) and (b), we show the mean inference accuracy $\mia$ for lag-1 correlation with a correction for confounding factors (\confoundingfactor), lag-1 correlation with a correction for reverse causation (\reversecausation), Granger causality (\grangercausality), and transfer entropy (\transferentropy) with increasing anticipated maximum transmission lag $\estmaxlag$ on edges. In (a), the actual maximum transmission lag $\maxlag$ matches $\estmaxlag$. In (b), we show results for constant $\maxlag$. In (c)--(f), we show results for varying $\maxlag$ and $\estmaxlag$ independently. (For values of  other parameters, see \cite{dpars}.).}
\label{fig:A3}
\end{figure}

Our comparisons of mean accuracy for network inference from dynamics with and without transmission lags on edges in Fig.\,\ref{fig:A1} suggest that for PEMs that do not account for transmission lags on edges (e.g., \laggedcorrelation{} and \ouinference{}), the mean inference accuracy decreases as the maximum transmission lag $\maxlag$ increases. Most of our considered PEMs accounting for transmission lags on edges have a parameter that determines the maximum edge transmission lag that one assumes when inferring a network. Our proposed PEMs, \reversecausation{} and \confoundingfactor{}, include the assumed maximum transmission lag $\estmaxlag$ as a parameter (see Appendix \ref{sec:els-proposals2}). For \grangercausality{}, the parameter setting the assumed maximum transmission time is the maximum order $\hat p$ of the VAR($p$) model used for regressing the time-series data.  For \transferentropy{}, the maximum edge transmission time $\hat p$ is given by the parameter that sets the number of data points that one uses to compute transfer entropy. The assumed maximum transmission lag on edges for \grangercausality{} and \transferentropy{} is $\estmaxlag=\hat p-1$.

For the results shown in Figs.\,\ref{fig:A1} and \ref{fig:A2}, we chose the assumed maximum transmission lag $\estmaxlag$ for the network inference to be equal to the actual maximum transmission lag $\maxlag$ that we used in the simulation of the SDD model. To demonstrate how a mismatch of $\maxlag$ and $\estmaxlag$ can affect the accuracy of inferred networks, we show the mean inference accuracy for \reversecausation{}, \confoundingfactor{}, \grangercausality{}, and \transferentropy{} with independently varying $\maxlag$ and $\estmaxlag$ in Fig.\,\ref{fig:A3}. 

In Fig.\,\ref{fig:A3}\,(a), we show results for increasing $\maxlag$ when the assumed maximum transmission lag on edges is equal to the actual maximum transmission lag on edges. We find that an increased transmission lag on edges does not have a negative effect on mean inference accuracy with \reversecausation{}, \confoundingfactor{}, or \grangercausality{} as a PEM when setting $\estmaxlag=\maxlag$. The mean inference accuracy for \transferentropy{} decreases with increasing $\maxlag$ even if one sets $\estmaxlag=\maxlag$. 

We show results for mean inference accuracy for a mismatch of the assumed and actual lag parameter (i.e., $\estmaxlag\neq\maxlag$) in Fig.\,\ref{fig:A3}\,(b). For the four considered PEMs, one obtains the best inference results when setting $\estmaxlag$ equal to or larger than $\maxlag$. When one underestimates the actual maximum edge transmission lag by setting $\estmaxlag<\maxlag$, the mean inference accuracy for all four PEMs tends to be much lower than when the assumed maximum edge transmission lag and the actual maximum edge transmission lag match (i.e., $\estmaxlag=\maxlag$). For all four PEMs with a parameter that sets $\estmaxlag$, overestimating the maximum edge transmission lag (i.e., $\estmaxlag>\maxlag$) tends to have a small negative effect or no effect on the mean inference accuracy. To demonstrate that these observations regarding a possible mismatch of $\estmaxlag$ and $\maxlag$ also hold for $\maxlag\neq 5$, we show heat maps of the mean inference accuracy for \reversecausation{}, \confoundingfactor{}, \grangercausality{}, and \transferentropy{} for varying $\maxlag$ and $\estmaxlag$  in Fig.\,\ref{fig:A3}\,(c)--(f).

\subsection{Inferring different network structures}\label{sec:resultsB}

\begin{figure}[!t]
\centering
\includegraphics[trim={0.48cm, 0.0cm, 0.1cm, 0.3cm}, clip,width=0.46\textwidth]{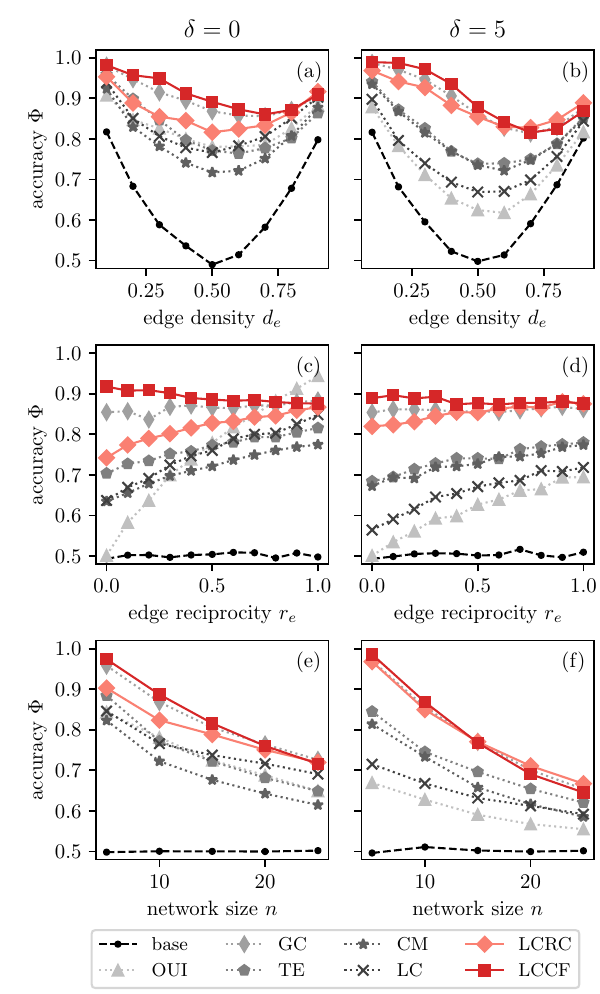}
\caption{{\bf Accuracy with varying structural properties of networks}. We compare inference results from using lag-1 correlation (\laggedcorrelation{}), transfer entropy (\transferentropy{}), linear Granger causality (\grangercausality{}), convergent cross mapping (\convergentcrossmapping{}), Ornstein--Uhlenbeck inference (OUI), and our proposed methods \confoundingfactor{} and \reversecausation{} (see Eqs.\,\eqref{eq:lcf} and \eqref{eq:lrc}) to infer networks that we sampled from a directed $G(n,m)$ random-graph model. The baseline ``base'' corresponds to inferring a network by selecting the correct number of edges uniformly at random. 
In (a) and (b), we vary the edge density $\density$. In (c) and (d), we vary the edge reciprocity $\reciprocity$. In (e) and (f), we vary the network size $\netsize$.
 In (a), (c), and (e), we show the results for dynamics with no transmission lag on edges (i.e., $\maxlag=0$). In (b), (d), and (f), we show results for dynamics with a maximum transmission lag $\maxlag=5$ on edges. 
(For default parameters, see \cite{dpars}.) Each data point corresponds to the mean over the accuracy of inferred networks for 100 ground-truth networks.}
\label{fig:B1}
\end{figure}

For the results that we presented in Subsection \ref{sec:resultsA}, we simulated dynamical systems on networks that are realizations of a $G(n,m)$ random-graph model with $\netsize=10$ nodes, an edge density $\density=0.5$, and an edge reciprocity of $\reciprocity=0.5$. In the current subsection, we investigate how changes in structural properties of ground-truth networks affect the mean inference accuracy for different PEMs. To that end, we show the mean inference accuracy for several PEMs as we vary the edge density $\density$, the edge reciprocity $\reciprocity$, and the network size $\netsize$ of ground-truth networks in Fig.\,\ref{fig:B1}.

\subsubsection{Edge density}
In Fig.\,\ref{fig:B1}\,(a), we show our results for mean inference accuracy as one varies edge density. If the number of edges is known, the mean baseline accuracy $\mia_\textrm{base}$ (i.e., the accuracy achieved by guessing the positions of edges in a network uniformly at random) is close to 1 for very sparse and very dense networks and $\mia_{base}\approx0.5$ for networks with edge density $\density=0.5$. For all PEMs that we consider in this paper, the mean inference accuracy changes with increasing edge density in a similar way to $\mia_{base}$. The mean inference accuracy for all considered PEMs is high when networks are very sparse or very dense and low when networks have an edge density of $\density\approx0.5$. In contrast to $\mia_{base}$ however, the mean inference accuracy for the considered PEMs is not symmetric about $\density=0.5$. Instead, all considered PEMs tend to infer networks with a small edge density $\density=d$ with higher accuracy than networks with a correspondingly large edge density $1-d$. This asymmetry is more pronounced for \confoundingfactor{} and \grangercausality{} than for the other considered PEMs and becomes stronger for \confoundingfactor{}, \reversecausation{}, and \grangercausality{} when the SDD model includes transmission lags on edges (see Fig.\,\ref{fig:B1}\,(b)). 

For \confoundingfactor{} and \reversecausation{}, we hypothesize that this asymmetry is connected to the existence of many long process motifs in very dense networks. When setting a value for the correction factor $\alpha$ for \confoundingfactor{} and \reversecausation{} in Section \ref{sec:els-proposals1}, we aimed to nullify the contributions of short process motifs (i.e., the motifs (1,1) and (1,0)) to lag-1 covariance. We did not consider how the resulting correction terms affect the contributions of long process motifs because long process motifs tend to have very small contributions to lag-1 covariance per motif occurrence. However, networks that are very dense include many long process motifs whose cumulative contribution to lag-1 covariance can be large and thus can negatively affect the mean inference accuracy when inferring very dense networks with \confoundingfactor{} or \reversecausation{}.

\subsubsection{Edge reciprocity}\label{sec:resultsB-reci}
In Fig.\,\ref{fig:B1}\,(c) and (d), we show our results for mean inference accuracy as one varies edge reciprocity of ground-truth networks with no edge transmission lags and with edge transmission lags of up to 5 timesteps, respectively. We find for both cases that, with most of the considered PEMs, one can infer networks with high edge reciprocity more accurately than networks in which most edges are not reciprocated. The increase of mean inference accuracy with increasing edge reciprocity is most pronounced when using \ouinference{} as a PEM. This PEM can only infer undirected networks accurately; edges in networks inferred via \ouinference{} are always reciprocated. The only PEM for which we find that mean inference accuracy is larger for low edge reciprocity than for high edge reciprocity is \confoundingfactor{}. An possible explanation for this observation is that a process motif $(0,1)$ (i.e., a length-1 walk from $j$ to $i$) in an undirected network is always accompanied by a process motif $(1,0)$ (i.e., a length-1 walk from $i$ to $j$), which has a negative contribution to \confoundingfactor{} and thus makes the detection of an undirected edge via \confoundingfactor{} less likely than that of a directed edge. The high mean inference accuracy for \confoundingfactor{} at low edge reciprocities suggests an interesting use case for \confoundingfactor{} as a PEM for inferring networks that one knows to have a low edge reciprocity.

\subsubsection{Network size}

In Fig.\,\ref{fig:B1}\,(e) and (f), we show our results for mean inference accuracy as one varies the network size. Whether the dynamics include transmission lags on edges or not, the mean inference accuracy decreases for all considered PEMs with increasing network size.

In Appendix \ref{app:shootingstar}, we compare several inferred networks to their corresponding ground-truth networks. For a set of small networks, which we call ``shooting-star networks'', we find that the coexistence of (1) nodes with low degree and (2) nodes with high degree and small local clustering coefficient makes network inference with \reversecausation{} difficult.
This observation suggests a possible explanation for the decreasing mean inference accuracy with increasing network size: with increasing network size, the probability of nodes with high degree and small local clustering coefficient in the Erd\H{o}s--R\'enyi model increases, and the increase of this probability makes it harder to infer these networks accurately. 

Regardless of the cause of the low mean inference accuracy for large networks, our results point to a serious problem for network inference via PEMs. The PEMs that we have considered here include several PEMs that researchers have used to infer large real-world networks with hundreds of nodes from time-series data \cite{Millington2019, Novelli2019}. We find that the mean inference accuracy for these PEMs approaches $\mia_{base}$ with increasing network size for our simulated time-series data. It is already close to (i.e., within $10\%$ of) baseline accuracy for networks with 50 nodes or more. Inferring large networks accurately via PEMs thus seems infeasible in practice without improved PEMs or a carefully designed preprocessing routine for time-series data.

For our default parameter values \cite{dpars}, we find that \confoundingfactor{}, \reversecausation{}, and \grangercausality{} tend to lead to the highest mean inference accuracies. As we have seen in Section \ref{sec:resultsA}, a ranking of these three PEMs by mean inference accuracy depends on one's choice of the parameters $\ctime$, $\epsilon$, and $\Delta t$. We revisit the interplay between parameters of random-graph models and the parameters of the dynamics on networks in Subsection \ref{sec:tau-structure}.

\subsubsection{Other variations of network structure}\label{sec:randomgraphmodels}

\begin{figure}[!t]
\centering
\includegraphics[trim={0.05cm, 0.1cm, 0.1cm, 0.1cm}, clip,width=0.46\textwidth]{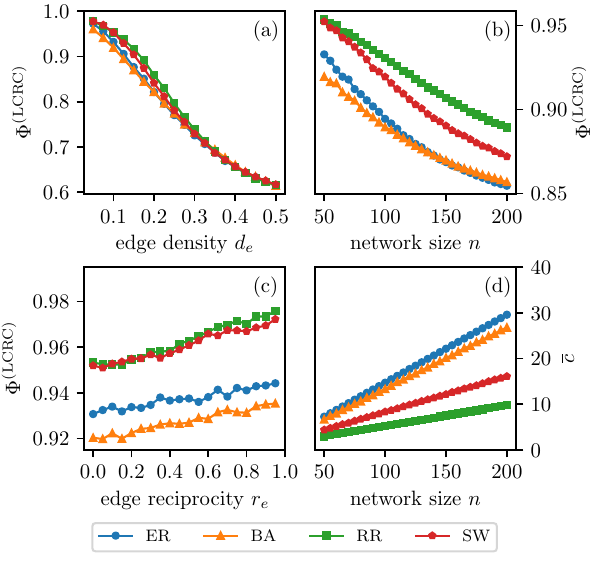}
\caption{{\bf Inference of realizations of graph models.} We show the mean accuracy $\mia$ of inferring networks via lag-1 correlation with a correction for reverse causation (\reversecausation{}) for variants of the Erd\H{o}s--Re\'enyi (ER) model, the Barab{\'a}si--Albert (BA) model, a regular-ring (RR) model, and the Watts--Strogatz small-world (SW) model. In (a), we vary the edge density $\density$. In (b), we vary the network size $\netsize$. In (c), we vary the edge reciprocity $\reciprocity$. In (d), we show the mean anti-clustering coefficient $\anticlust$ with increasing network size. We observe that the inference accuracy tends to be large when $\anticlust$ is small and vice versa.}
\label{fig:B2}
\end{figure}
To investigate how other aspects of the structure of ground-truth networks can affect mean inference accuracy, we compare the mean inference accuracy $\mia_{\textrm{\reversecausation{}}}$ for \reversecausation{} for four graph models: a regular ring, the Watts-Strogatz small-world model \cite{Watts1998}, the Barab{\'a}si-Albert model \cite{Barabasi1999}, and the Erd\H{o}s-R\'enyi model \cite{Newman2018}. We use model variants for which one can specify the network size, edge density, and edge reciprocity as model parameters (see implmentations in our accompanying code library \cite{code}).

In Fig.\,\ref{fig:B2}, we show the mean inference accuracy for \reversecausation{}, $\mia_{\textrm{\reversecausation{}}}$, for realizations of the four considered graph models. We find that the relationships between $\mia_{\textrm{\reversecausation{}}}$ and edge density $\density$, edge reciprocity $\reciprocity$, and network size $\netsize$ are qualitatively similar for realizations of all four considered graph models; the mean inference accuracy decreases with increasing edge density $\density\in(0,0.5]$ (see Fig.\,\ref{fig:B2}\,(a)) or increasing network size (see Fig.\,\ref{fig:B2}\,(b)), and it increases with increasing edge reciprocity (see Fig.\,\ref{fig:B2}\,(c)). 

For almost all parameter combinations that we considered, we obtain much higher $\mia_{\textrm{\reversecausation{}}}$ for the regular-ring model and the Watts--Strogatz model than for the Barab{\'a}si--Albert model and the Erd\H{o}s--R\'enyi model. A possible explanation for the differences in mean inference accuracy between the four graph models is that nodes with a high degree and small local clustering coefficient --- whose neighborhoods tend to be difficult to infer (see Appendix \ref{app:shootingstar}) --- are more frequent in realizations of the Erd\H{o}s--R\'enyi model and the Barab{\'a}si--Albert model than in realizations of the regular-ring model or the Watts-Strogatz model. 

To characterize nodes with a high degree and a small local clustering coefficient, we define the \textit{anti-clustering coefficient} of a node $i$ with degree $k_i$ and local clustering coefficient $c_i$ as
\begin{align}
    \overline{c}_i:=k_i(1-c_i)\,.\label{eq:anticlustering}
\end{align}
In Fig.\,\ref{fig:B2}\,(d), we show the mean anti-clustering coefficient $\overline{c}:=\frac{1}{n}\sum_i \overline{c}_i$ over all nodes $i$ in a network as we increase the network size. Comparing Fig.\,\ref{fig:B2}\,(b) with Fig.\,\ref{fig:B2}\,(d), we observe that the mean inference accuracy tends to be high when the mean anti-clustering coefficient is small (Spearman correlation $r_s=-0.93$ with $p<10^{-100}$).

\subsubsection{Interplay of the characteristic time of the stochastic delay-difference model and structural properties of networks}\label{sec:tau-structure}

\begin{figure}[!b]
\centering
\includegraphics[trim={0.0cm, 0.0cm, 0.2cm, 0.0cm}, clip,width=0.46\textwidth]{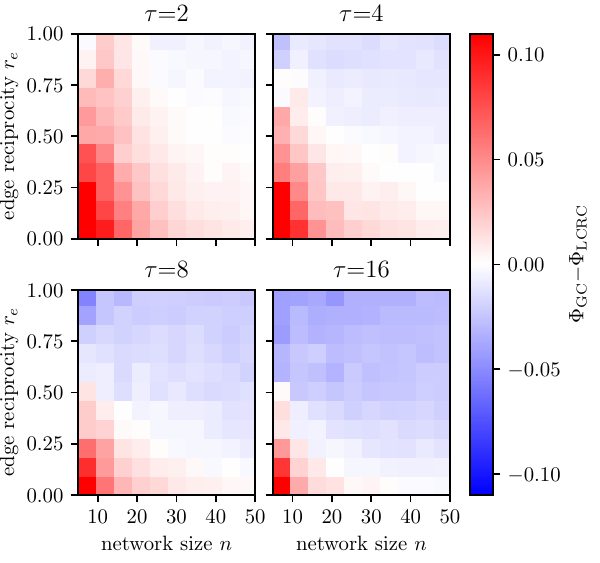}
\caption{{\bf Comparison of Granger causality and lag-1 correlation with a correction for reverse causation.} We show the difference of the mean accuracy $\mia_{\textrm{\grangercausality{}}}$ for inferring networks with Granger causality (\grangercausality) and the mean accuracy $\mia_{\textrm{\reversecausation{}}}$ for inferring networks with lag-1 correlation with a correction for reverse causation (\reversecausation{}). We vary the network size $\netsize$, the edge reciprocity $\reciprocity$, and the characteristic time $\ctime$. (For values of  other parameters, see \cite{dpars}.) Red pixels indicate that \grangercausality{} leads to higher mean accuracy than \reversecausation{}. Blue pixels indicate that \grangercausality{} leads to lower mean accuracy than \reversecausation{}.}
\label{fig:B3}
\end{figure}

In this subsection, we demonstrate that results on the ranking of PEMs based on their associated mean inference accuracy can depend on an interplay of parameters of the dynamics on a network and the structural properties of the network.

In Fig.\,\ref{fig:B3}, we show the difference $\mia_{\textrm{\grangercausality{}}}-\mia_{\textrm{\reversecausation{}}}$ of mean inference accuracy for \grangercausality{} and \reversecausation{} with varying network size and edge reciprocity for four different values of the characteristic time $\ctime$. We observe that $\mia_{\textrm{\grangercausality{}}}-\mia_{\textrm{\reversecausation{}}}$ decreases and even changes from positive to negative values as one increases the network size or edge reciprocity of ground-truth networks, or the characteristic time of the SDD model. When using mean inference accuracy as the only criterion for deciding which PEM to use for network inference, one thus has to take into account structural properties of the networks that one wants to infer as well as parameters of the dynamics that give rise to the observed time-series data.

\subsection{Effect of noisy dynamics and noisy measurements}\label{sec:resultsC}

\begin{figure}[!t]
\centering
\includegraphics[trim={0.48cm, 0.0cm, 0.0cm, 0.0cm}, clip,width=0.46\textwidth]{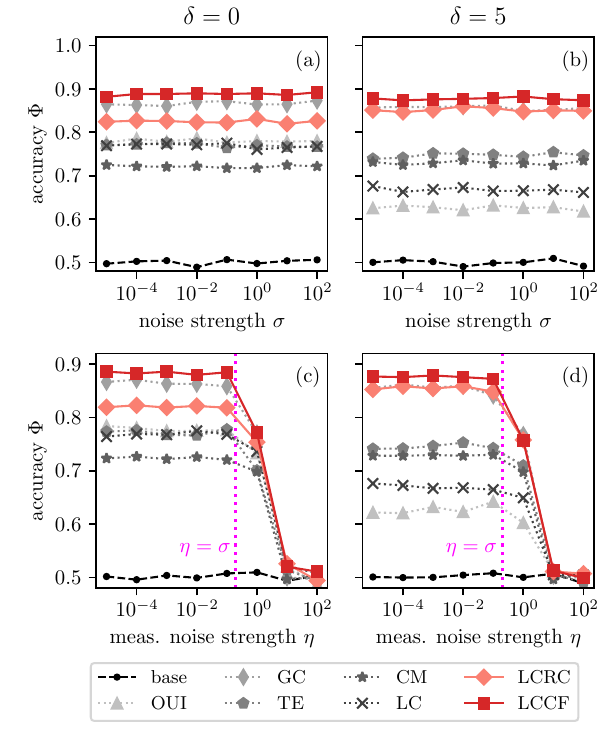}
\caption{{\bf Accuracy with varying noise strengths.} We compare inference results from using several pairwise edge measures (PEMs) to infer networks that we sampled from a directed $G(n,m)$ random-graph model. In (a) and (b), we vary the strength $\noise$ of  system noise. In (c) and (d), we vary the strength $\mnoise$ of measurement noise. A vertical dotted magenta line indicates where $\mnoise=\noise$.
 In (a) and (c), we show the results for dynamics with no transmission lag on edges (i.e., $\maxlag=0$). In (b), and (d), we show results for dynamics with a maximum transmission lag $\maxlag=5$ on edges. 
 For values of  other parameters, see \cite{dpars}. For PEM acronyms, see caption of Fig.\,\ref{fig:A1}.}
\label{fig:C1}
\end{figure}

In this subsection, we study the effect of noise on mean inference accuracy. We consider noise in the dynamics and noise in the measurement process. In the SDD model, we model noise in the dynamical system as additive Gaussian white noise in the delay-difference equation (see Eq.\,\eqref{eq:doup1}). The parameter that tunes the strength of this system noise is $\noise$. We model measurement noise as additive Gaussian white noise added to each observation in a time-series data set. The parameter $\mnoise$ tunes the strength of the measurement noise. 

In Fig.\,\ref{fig:C1}, we show our results for mean inference accuracy as one varies the strength $\noise$ of the system noise and the strength $\mnoise$ of the measurement noise. In panels (a) and (b), we show the mean inference accuracy for several PEMs as one increases the strength of system noise for the SDD model with maximum edge transmission lag on edges of $\maxlag=0$ and $\maxlag=5$, respectively. For all considered PEMs, our results indicate that the strength of system noise does not affect mean inference accuracy. In panels (c) and (d), we show our results for increasing the strength of measurement noise. We observe that the mean inference accuracy for all considered PEMs is unaffected by measurement noise when $\mnoise\lesssim\noise$. When the strength of the measurement noise exceeds the strength of the system noise, it leads to a decrease in mean inference accuracy for all considered PEMs.

\subsection{Nonstationarity and data efficiency}\label{sec:resultsD}

\begin{figure*}[!t]
\centering
\includegraphics[trim={0.0cm, 0.0cm, 0.0cm, 0.0cm}, clip,width=1\textwidth]{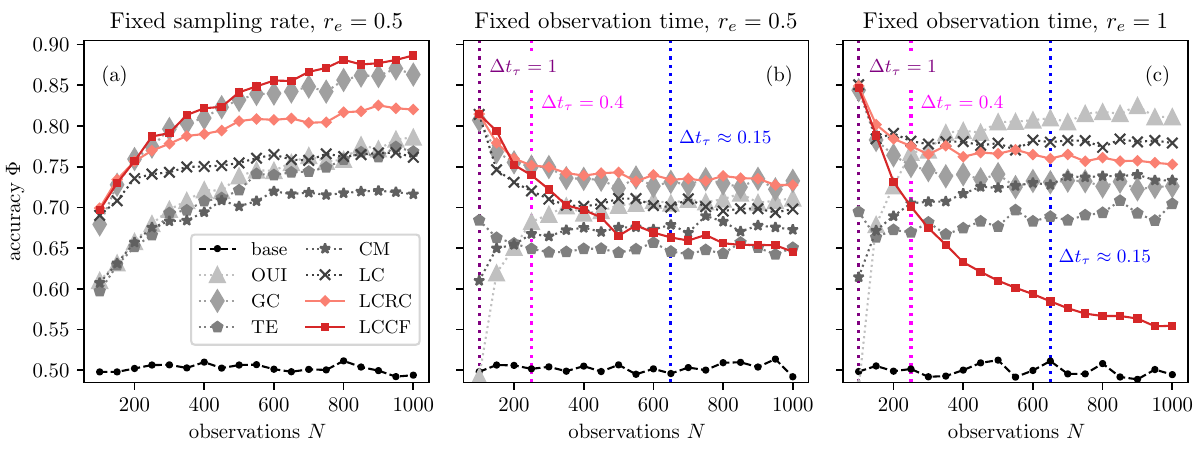}
\caption{{\bf Accuracy with varying the number of observations.} We compare inference results from using several pairwise edge measures (PEMs) to infer networks that we sampled from a $G(n,m)$ random-graph model. In (a), we vary the number $N$ of observations while keeping the sampling period $\Delta t$ fixed and thus varying the observation time $T$. In (b) and (c), we vary the $N$ while keeping $T$ fixed and thus varying $\Delta t$. Dotted vertical lines indicate where $N$ is such that $\Delta t_\ctime=2$, $\Delta t_\ctime=1$ and where $\Delta t_\ctime=\frac{1}{2}$. We show results for directed networks in (b) and results for undirected networks in (c). The labels of the dashed vertical lines at $\numsamples=100$, $\numsamples=250$, and $\numsamples=650$ indicate the value of $\Delta t_\ctime$ corresponding the respective values of $\numsamples$. All results are for networks with a maximum transmission lag $\maxlag=0$ on edges. For values of  other parameters, see \cite{dpars}. For PEM acronyms, see caption of Fig.\,\ref{fig:A1}.
}
\label{fig:D1}
\end{figure*}

When calculating \laggedcorrelation{}, \grangercausality{}, \transferentropy{}, \ouinference, \reversecausation{}, and \confoundingfactor{}, one estimates the steady-state covariance matrices or transfer entropy of a stationary stochastic process from time-series data. To ensure that the sample covariance matrix of a set of time-series data is a good approximation of the steady-state covariance matrix of the underlying stochastic process, one needs a large number of independent observations. What happens when this is not the case?

In Fig.\,\ref{fig:D1}, we investigate the effect of the number $\numsamples$ of samples on the mean inference accuracy.
As one increases the number of observations, one either also increases the observation time $T$ while keeping the sampling rate $1/\Delta t$ fixed (see Fig.\,\ref{fig:D1}\,(a)) or one increases the sampling rate while keeping $T$ fixed (see Fig.\,\ref{fig:D1}\,(b--c)) . 

\subsubsection{Fixed sampling rate vs. fixed observation time}

When one increases $\numsamples$ and $T$ while keeping $1/{\Delta t_\ctime}$ fixed, one successively includes more and more independent observations in the data set. This increase of independent observations tends to lead to an increase of the mean inference accuracy for all considered PEMs (see Fig.\,\ref{fig:D1}\,(a)).

When one increases $\numsamples$ while keeping $T$ fixed, one adds more samples to the data set but increases the statistical dependence between subsequent data points. The the ratio $\Delta t_\ctime:=\Delta t/\ctime$ changes, and this change can have a strong effect on mean inference accuracy $\mia$ (see Fig.\,\ref{fig:D1}\,(b--c)).
At $\Delta t_\ctime=1$, the SDD model is equivalent to a VAR model, and we thus expect PEMs that perform well on VAR models (i.e., \grangercausality{}, \transferentropy{}, \laggedcorrelation{}) to achieve their highest mean inference accuracies when $\Delta t_\ctime=1$. As $\Delta t_\ctime$ decreases, the SDD model becomes more and more similar to the OUP, and we thus expect the mean inference accuracy of PEMs for continuous-time processes (i.e., \ouinference{} and \convergentcrossmapping{}) to increase. Our results in Fig.\,\ref{fig:D1}\,(b--c) are consistent with these hypotheses.

\subsubsection{Varying $\Delta t_\ctime$ in directed and undirected networks}
For networks with edge reciprocity $\reciprocity=0.5$, network inference via \grangercausality{} tends to be more accurate than network inference via \ouinference{}, even for large $\numsamples$ and small $\Delta t_\ctime$. This is because it is impossible to accurately infer directed networks via \ouinference{}, because \ouinference{} is symmetric (i.e., $\named{\score}{\ouinference}_{i,j}=\named{\score}{\ouinference}_{j,i}$). For undirected networks (i.e., $\reciprocity=1$), the mean inference accuracy $\named{\mia}{\ouinference}$ is close to $\named{\mia}{base}$ when $\Delta t_\ctime=1$, but it rapidly increases in with increasing $N$ and decreasing  $\Delta t_\ctime$. At $N\approx 250$ (i.e., $\Delta t_\ctime\approx0.4$), we observe similar mean inference accuracy for \ouinference{} and \grangercausality{}. At $N\approx 650$ (i.e., $\Delta t_\ctime\approx0.15$), the mean inference accuracy of \convergentcrossmapping{} is comparable to \grangercausality{}.

In accordance with results in previous subsections, we find that (i) \confoundingfactor{} and \reversecausation{} infer networks with similar or higher accuracy than all other considered PEMs when $\Delta t_\ctime\in[0.5, 1]$ (see Section \ref{sec:resultsA-dt}), (ii) \reversecausation{} tends to infer directed networks with higher accuracy than the other considered PEMs when $\Delta t_\ctime>0.5$ (see Section \ref{sec:resultsA-tau}), and (iii) \confoundingfactor{} performs much better on directed networks than on undirected networks (see Section \ref{sec:resultsB-reci}).

Our results demonstrate that collecting more data on a stochastic process does not necessarily improve the accuracy of inferred networks. For many PEMs, including our proposed PEMs \reversecausation{} and \confoundingfactor{}, one obtains the most accurately inferred networks when $\Delta t \approx \ctime$. We conclude that it is important to consider the characteristic time of a stochastic process on a network when setting the sampling rate to achieve a high accuracy of inferred networks. 

\subsection{Computation time}\label{sec:resultsE}

\begin{figure}[!t]
\centering
\includegraphics[trim={0.2cm, 0.4cm, 0.08cm, 0.0cm}, clip,width=0.46\textwidth]{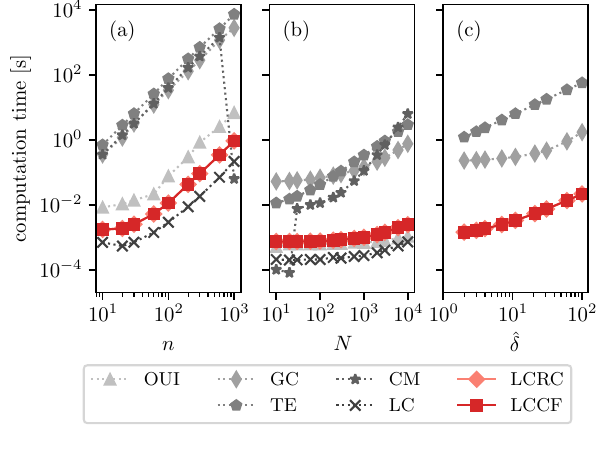}
\caption{{\bf Comparison of computation time for pairwise edge measures (PEMs).} In (a), we vary the network size $\netsize$. In (b), we vary the number $\numsamples$ of samples. In (c), we vary the anticipated maximum transmission lag $\estmaxlag$ on edges. (For default parameters and acronyms for PEMs, see caption of Fig.\,\ref{fig:A1}.) Each data point corresponds to the mean over the computation time for 10 ground-truth networks.}
\label{fig:E1}
\end{figure}

In Subsections \ref{sec:resultsA}--\ref{sec:resultsD}, we investigated how various parameters of graph models and stochastic processes on networks affect the mean inference accuracy for several PEMs. In addition to accuracy, the computational costs associated with performing network inference may be a relevant consideration when selecting a PEM. In Fig.\,\ref{fig:E1}, we show our results for the mean computation time for inferring a network via one of several PEMs as one varies the size of the network $\netsize$, 
the number of samples in a time-series data set $\numsamples$, and the anticipated maximum edge transmission lag $\estmaxlag$.

Our results for increasing the network size (see Fig.\,\ref{fig:E1}\,(a))  demonstrate that the mean computation time increases fast with increasing network size for all considered PEMs. However, one can group the seven PEMs that we considered into two groups --- one of computationally expensive PEMs and one of computationally inexpensive PEMs. The group of computationally expensive PEMs includes \grangercausality{}, \transferentropy{}, and \convergentcrossmapping{}. The group of computationally inexpensive PEMs includes \laggedcorrelation{}, \ouinference, \confoundingfactor{}, and \reversecausation{}. For all considered network sizes, the mean computation time associated with any of the computationally expensive PEMs is at least 100 times longer than the mean computation time associated with any of the computationally inexpensive PEMs. The only exception to this statement is the mean computation time associated with \convergentcrossmapping{} at $\netsize=1000$. At this network size, a time-series data set with $\numsamples=1000$ observations is too small to compute \convergentcrossmapping{}. In that case, the algorithm stops without returning an inferred network.

In Fig.\,\ref{fig:E1}\,(b), we show results for increasing the number $\numsamples$ of observations. We find that the mean computation time for all considered PEMs increases with $\numsamples$. The mean computation time for \grangercausality{}, \transferentropy{}, and \convergentcrossmapping{} increases more strongly with $\numsamples$ than the mean computation time for the other considered PEMs. Low mean computation times for \convergentcrossmapping{} for $N\lesssim50$ occur because the algorithm aborts due to an insufficient number of observations.

For PEMs that include a parameter to set the anticipated maximum edge transmission lag $\estmaxlag$, we show how the mean computation time changes with $\estmaxlag$ in Fig.\,\ref{fig:E1}\,(c). We again find that we can group the considered PEMs into a group of computationally expensive PEMs, which includes \grangercausality{} and \transferentropy{}, and a group of computationally inexpensive PEMs, which includes \confoundingfactor{} and \reversecausation{}.

When comparing the mean inference accuracy for several PEMs in Subsections \ref{sec:resultsA}--\ref{sec:resultsD}, we consistently obatined the most accurate network inferences for many parameter pairs with \grangercausality{}, \confoundingfactor{}, or \reversecausation{}. When computational costs are a relevant consideration for selecting a network inference method, we believe that \confoundingfactor{} and \reversecausation{} offer a very good trade-off between highly accurate network inference and low computational costs. We do not find a comparably good combination of high mean inference accuracy and low computation costs for any other PEM that we considered in our study.

\section{Discussion}\label{sec:discussion}

Inferring networks from time-series data is a challenging problem with many applications. Many researchers have used vector-autoregressive (VAR) models to fit time-series data and infer networks from it via pairwise edge measures (PEMs), such as lag-1 correlation, Granger causality, and transfer entropy.  We proposed a stochastic delay-difference (SDD) model (see Eq.\,\eqref{eq:doup1})  that is a generalization of a VAR model (see Eq.\,\eqref{eq:varp}) and that allows for a mismatch of the characteristic time $\ctime$ of the dynamics and the sampling rate $1/\Delta t$. We then derived the process motifs for covariance and lagged covariance for the SDD model, and we used the contributions of process motifs to covariance and lagged covariance to formulate a family of PEMs that are linear combinations of the sample correlation matrix and the lag-1 correlation matrix of a time-series data set. From this family of PEMs, we selected two members that have a particularly strong theoretical motivation.

Our first proposed PEM is lag-1 correlation with a correction for confounding factors (\confoundingfactor{}, see Eq.\,\eqref{eq:lcf}), which one can use to infer networks with high accuracy from mildly autocorrelated time-series data (i.e., $0.3\lesssim\Delta t/\ctime\leq 1$). Our second proposed PEM is lag-1 correlation with a correction for reverse causation (\reversecausation{}, see Eq.\,\eqref{eq:lrc}), which leads to accurate network-inference results for strongly autocorrelated time-series data (i.e., $\Delta t/\ctime\lesssim0.3$). Importantly, one can compute both of our proposed PEMs more easily and with less computational resources than many widely-used PEMs, while obtaining similar or better inference results than the widely-used PEMs for many different network structures and for large regions of the parameter space of the SDD model. When considering inference accuracy and computational costs in the selection of a PEM for network inference, we suggest that our proposed PEMs present a better trade-off between high inference accuracy and low computational cost for the inference of networks from linear stochastic dynamics than any other PEM that we are currently aware of.

We anticipate the strong performance of our proposed PEMs is limited to the inference of linear stochastic dynamics with additive Gaussian white noise. Non-linear dynamics and dynamics with system noise that is not additive or not Gaussian are not well modelled by our SDD model, which is the theoretical base of our proposed PEMs. We do not anticipate, however, that linear Granger causality leads to much better inference results than our proposed PEMs in either of these settings. The theoretical base for linear Granger causality is a VAR model (see Eq.\,\eqref{eq:varp}) and any process that is not well modelled by our SDD model is also not well modelled by this VAR model.

In future work, we would like to conduct a systematic comparison of network inference via our proposed PEMs to alternative approaches to causal inference from time-series data sets with strong autocorrelation. For example, many researchers have used a combination of preprocessing strategies (e.g. de-trending \cite{Kang1985, Hussain2017} and subsampling \cite{Faes2017, Barnett2017}) with PEMs to circumvent problems arising from strong autocorrelation in stochastic processes. A possible future direction for extending the applicability of our PEMs would be to develop significance tests for \confoundingfactor{} and \reversecausation{}. Such tests would allow researchers to use \confoundingfactor{} and \reversecausation{} for network inference even when the number of edges in a network is not known a priori. Finally, our results demonstrated that the mean inference accuracy for many PEMs decreases with increasing network size. While our current paper includes results that suggest that a cause of the decline of mean inference accuracy with network size is connected to an increase of a network's mean anticlustering coefficient (see Section \ref{sec:resultsB}), we anticipate that further research on the challenges posed by large networks to inference methods is a particularly interesting avenue of future research. One can compute our proposed PEMs at low computational cost for very large networks with hundreds or thousands of nodes, but we do not expect inference results for such large networks to be  more accurate than guessing edge positions uniformly at random. An improved understanding of the problems that large network sizes pose for network inference via PEMs has the potential to lead to improved PEMs that can infer large networks accurately with low computational cost.

\section*{Acknowledgements}
ACS would like to thank Kameron Dekker Harris, Joseph Lizier, and Leonardo Novelli, Ezekile Williams, and the MILA Neural-AI Reading Group for helpful discussions. BWB and ACS were supported by the Air Force Office of Scientific Research MURI FA9550-19-1-0386 (BWB). ACS received additional support from the eScience Institute at the University of Washington. SMI was supported by the undergraduate research fellowship program of the Weill Neurohub.

\appendix

\section{Derivation of the covariance matrix for the stochastic difference equation}\label{app:sigma-sol} 

Using the binomial theorem, one can express the $k$-th power of $\prop$ as
\begin{align}
    \prop^k = \sum_{\ell=0}^k \binom{k}{\ell} (1-\Delta t_\ctime)^{k-\ell}(\Delta t_\ctime \epsilon \admat{1})^{\ell}\,.
\end{align}
One can thus write 
\begin{align}
    \prop^k\left(\prop^T\right)^k = \sum_{\ell_B=0}^k \sum_{\ell_F=0}^k \TFA_{k,\ell_B,\ell_F}
\end{align}
for which we define the 3-index sequence,
\begin{align}
    \TFA_{k,\ell_B,\ell_F}
    &:= 
    \TFB_{k,\ell_B,\ell_F} \TFC_{\ell_B,\ell_F}\,,
\end{align}
which has a scalar-valued part,
\begin{align}
    \TFB_{k,\ell_B,\ell_F}
    &:= 
    \binom{k}{\ell_B}\binom{k}{\ell_F}
    (1-\Delta t_\ctime)^{2k-(\ell_B+\ell_F)}
    \Delta t_\ctime^{\ell_B+\ell_F}
    \,,
\end{align}
which depends on $k$ and $\Delta t_\ctime$, and a matrix-valued part
\begin{align}
    \TFC_{\ell_B,\ell_F}
    &:=\epsilon^{\ell_B+\ell_F}\left(\admat{1}\right)^{\ell_F}\left(\left(\admat{1}\right)^T\right)^{\ell_B}\,,\nonumber
\end{align}
which is independent of $k$ and $\Delta t_\ctime$.
The discrete-time Lyapunov equation Eq.\,\eqref{eq:covmat-prob} has the solution \cite[p.~108]{Fairman1998}
\begin{align}
    \ccovmat{0} &= \frac{\noise^2\Delta t}{\netsize}\sum_{k=0}^\infty \prop^k\left(\prop^T\right)^k\nonumber\\
    &= \frac{\noise^2\Delta t}{\netsize}\sum_{k=0}^\infty\sum_{\ell_B=0}^k \sum_{\ell_F=0}^k\TFA_{k,\ell_B,\ell_F}\label{eq:triplesum}
\end{align}
When all eigenvalues of $\epsilon\admat{1}$ have absolute values less than 1, the 3-index sum in Eq.\,\eqref{eq:triplesum} is absolutely convergent, and one can thus apply Fubini's theorem for infinite series and exchange the summation indices, such that, 
\begin{align}
    \ccovmat{0} &= \frac{\noise^2\Delta t}{\netsize}\sum_{\ell_B=0}^\infty \sum_{k=\ell_B}^\infty\sum_{\ell_F=0}^k \TFA_{k,\ell_B,\ell_F}\nonumber\\
    &= \frac{\noise^2\Delta t}{\netsize}\sum_{\ell_B=0}^\infty \sum_{\ell_F=0}^\infty \sum_{k=\max\{\ell_B,\ell_F\}}^\infty\TFA_{k,\ell_B,\ell_F}\nonumber\\
    &= \frac{\noise^2\Delta t}{\netsize}\sum_{\ell_B=0}^\infty \sum_{\ell_F=0}^\infty \sum_{k=0}^\infty\TFA_{k+\max\{\ell_B,\ell_F\},\ell_B,\ell_F}\,,\nonumber\\
    &= \frac{\ctime\noise^2}{\netsize}\sum_{\ell_B=0}^\infty \sum_{\ell_F=0}^\infty \TFC_{\ell_B,\ell_F} \sum_{k=0}^\infty\TFB_{k+\lmax,\ell_B,\ell_F}\Delta t_\ctime\,,
    \label{eq:sumswitch}
\end{align}
where we use the shorthand notation $\lmax:=\max\{\ell_B,\ell_F\}$.
To simplify the infinite sum over $k$ in Eq.\,\eqref{eq:sumswitch}, we introduce the shorthand notations 
$\lmin:=\min\{\ell_B,\ell_F\}$ and $\ldiff:=\lmax-\lmin$, such that,
\begin{widetext}
\begin{align}
    \sum_{k=0}^\infty\TFB_{k+\lmax,\ell_B,\ell_F}\Delta t_\ctime &= \sum_{k=0}^\infty\binom{k+\lmax}{\lmax}\binom{k+\lmax}{\lmin}(1-\Delta t_\ctime)^{2k+2\lmax-(\lmax+\lmin)}\Delta t_\ctime^{\lmax+\lmin+1}\,.\label{eq:almost2f1-0}
\end{align}
Using the relationship,
\begin{align}
\binom{k+\lmax}{\lmin}\binom{\lmax}{\ldiff}^{-1}=\binom{k+\lmax}{\lmax}\binom{k+\ldiff}{\ldiff}^{-1}\,,
\end{align}
one can write Eq.\,\eqref{eq:almost2f1-0} as
\begin{align}
    \sum_{k=0}^\infty\TFB_{k+\lmax,\ell_B,\ell_F}\Delta t_\ctime &= \Delta t_\ctime^{\lmax+\lmin+1}(1-\Delta t_\ctime)^{\lmax-\lmin}\binom{\lmax}{\ldiff}\sum_{k=0}^\infty\frac{\displaystyle\binom{k+\lmax}{\lmax}\binom{k+\lmax}{\lmax}}{\displaystyle\binom{k+\ldiff}{\ldiff}}(1-\Delta t_\ctime)^{2k}\,.
   \label{eq:almost2f1-1}
\end{align}
For processes with $\Delta t_\ctime\in(0,1)$), the infinite sum over $k$ in Eq.\,\eqref{eq:almost2f1-1} defines a hypergeometric function $_2F_1\hspace{-1mm}\left(\lmax+1,\lmax+1;\ldiff+1 | (1-\Delta t_\ctime)^2\right)$. One thus obtains
\begin{align}
    \sum_{k=0}^\infty\TFB_{k+\lmax,\ell_B,\ell_F}\Delta t_\ctime 
    &=\Delta t_\ctime^{\lmax+\lmin+1}(1-\Delta t_\ctime)^{\lmax-\lmin}\binom{\lmax}{\ldiff} \hypergeometric{\lmax+1,\,\,\lmax+1}{\ldiff+1}
    {(1-\Delta t_\ctime)^2}\\
    &=\mcf_{\ell_B,\ell_F}(\Delta t_\ctime)\,.\label{eq:finally2f1}
\end{align}
Using Eq.\,\eqref{eq:finally2f1} to substitute the infinite sum over $k$ in Eq.\,\eqref{eq:sumswitch}, one obtains Eq.\,\eqref{eq:covmat-sol}.

\section{Covariance matrix of the stochastic difference equation in the limit $\Delta t_\ctime\rightarrow 0$}\label{app:limit0}
For real numbers $z\neq 1$ \footnote{The Euler transformation holds for complex numbers $z$ with nonzero imaginary parts, too. We state the Euler transformation for real numbers $z$, because we use hypergeometric functions with real arguments $z$ throughout our paper.}, the hypergeometric function $_2F_1$ satisfies the Euler transformation \cite[p.~390]{NIST},
\begin{align}
    \hypergeometric{a,\, b}{c}{z}
 = (1-z)^{c-a-b} \hypergeometric{c-a,\, c-b}{c}{z}\,.\label{eq:euler}
\end{align}
At $z=1$, the hypergeometric function satisfies Gauss's hypergeometric theorem \cite[p.~387]{NIST}, which states that for $c-a-b>0$,
\begin{align}
    \hypergeometric{a,\, b}{c}{1}
     = \frac{\Gamma(c)\Gamma(c-a-b)}{\Gamma(c-a)\Gamma(c-b)}\,, \label{eq:Gauss} 
\end{align}
where $\Gamma$ is the Gamma function.
One can use the identities in Eqs.\,\eqref{eq:euler} and \eqref{eq:Gauss} to derive the limit, 
\begin{align}
   \lim_{z\rightarrow 0}\left[z^{p+q+1} \hypergeometric{\maxval{p,q},\,\, \maxval{p,q}}{|p-q|+1}{(1-z)^2}\right]
 = 2^{-(p+q+1)}\frac{\displaystyle\binom{p+q}{\maxval{p,q}-1}}{\displaystyle\binom{\maxval{p,q}-1}{|p-q|}}\,.\label{eq:limit_lemma}
\end{align}
To derive Eq.\,\eqref{eq:limit_lemma}, we start with the left-hand side of Eq.\,\eqref{eq:limit_lemma}, to which we apply the Euler transformation Eq.\,\eqref{eq:euler} to obtain
\begin{align}
    \lim_{z\rightarrow 0} \left[z^{p+q+1} \hypergeometric{\maxval{p,q},\,\, \maxval{p,q}}{|p-q|+1}{(1-z)^2}\right]
    &= \lim_{z\rightarrow 0}\Bigg[z^{p+q+1}(1-(1-z)^2)^{|p-q|+1-2\maxval{p,q}} \\
    &\phantom{=\lim_{z\rightarrow 0}\Bigg[}\left.\times\hypergeometric{|p-q|+1-\maxval{p,q},\, |p-q|+1-\maxval{p,q}}{|p-q|+1}{(1-z)^2}\right]\\
    &= \lim_{z\rightarrow 0}\left[(2-z)^{-(p+q+1)} \hypergeometric{-\min\{p,q\},\, -\min\{p,q\}}{|p-q|+1}{(1-z)^2}\right]\,.\label{eq:limit_lemma0}
\end{align}
For non-negative integers $p$ and $q$, the expression $p+q+1$ is greater than $0$. One can thus take the limit $z\rightarrow 0$ in Eq.\,\eqref{eq:limit_lemma0} and apply Gauss's hypergeometric theorem (see Eq.\,\eqref{eq:Gauss}) to obtain
\begin{align}
    \lim_{z\rightarrow 0} \left[z^{p+q+1} \hypergeometric{\maxval{p,q},\,\, \maxval{p,q}}{|p-q|+1}{(1-z)^2}\right]
    &=2^{-(p+q+1)}\frac{\Gamma(|p-q|+1)\Gamma(|p-q|+2\min\{p,q\}+1)}{\Gamma(|p-q|+\min\{p,q\}+1)\Gamma(|p-q|+\min\{p,q\}+1)}\\
    &=2^{-(p+q+1)}\frac{\Gamma(|p-q|+1)\Gamma(p+q+1)}{\Gamma(\max\{p,q\}+1)\Gamma(\max\{p,q\}+1)}\\
    &=2^{-(p+q+1)}\frac{\Gamma(|p-q|+1)\Gamma(p+q+1)}{\Gamma(\maxval{p,q})\Gamma(\maxval{p,q})}\,.\label{eq:limit_lemma1}
\end{align}
All arguments of Gamma functions in Eq.\,\eqref{eq:limit_lemma1} are positive integers. One can thus write
\begin{align}
    \lim_{z\rightarrow 0} \left[z^{p+q+1} \hypergeometric{\maxval{p,q},\,\, \maxval{p,q}}{|p-q|+1}{(1-z)^2}\right]
    &=2^{-(p+q+1)}\frac{(|p-q|)!(p+q)!}{(\maxval{p,q}-1)!(\maxval{p,q}-1)!}\\
    &=2^{-(p+q+1)}\frac{(p+q)!}{(\maxval{p,q}-1)!(\min\{p,q\})!}\frac{(\min\{p,q\})!(|p-q|)!}{(\maxval{p,q}-1)!}\\
    &=2^{-(p+q+1)}\frac{\displaystyle\binom{p+q}{\maxval{p,q}-1}}{\displaystyle\binom{\maxval{p,q}-1}{|p-q|}}\,,\label{eq:limit_lemma2}
\end{align}
which proves Eq.\,\eqref{eq:limit_lemma}.

One can use Eq.\,\eqref{eq:limit_lemma} to derive the limit 
\begin{align}
\lim_{\Delta t_\ctime\rightarrow 0} \con{0}_{\ell_B,\ell_F} &= \frac{\ctime\noise^2}{\netsize}\epsilon^{\ell_B+\ell_F} \lim_{\Delta t_\ctime\rightarrow 0}\mcf_{\ell_B,\ell_F}(\Delta t_\ctime)
= \frac{\ctime\noise^2}{2^{\ell_B+\ell_F+1}\netsize}\epsilon^{\ell_B+\ell_F}\binom{\ell_B+\ell_F}{\ell_F}\,.\label{eq:limit1}
\end{align}

\end{widetext}

Substituting $\ell_F$ by $\ell$ and $\ell_B$ by $L-\ell_F$ in Eq.\,\eqref{eq:limit1}, one can see that the limit of the contributions $\con{0}_{\ell_B,\ell_F}$ of process motifs to the steady-state covariances in the SDD model is identical to the contributions of process motifs to the steady-state covariances in the OUP (see Eq.\,\eqref{eq:covmat-contributions-ou}) in the limit $\Delta t_\ctime\rightarrow 0$.

\section{Estimation of the inverse characteristic time}\label{sec:theta_est}
To compute the values of the PEMs that we proposed in Section \ref{sec:els-proposals1}, one needs to know the value  
of $\ctimeinv$. For the case that this parameter is not known, we provide an estimator $\estctimeinv$ that one can compute from the sample correlation matrix and the lag-1 sample correlation matrix of the time-series data. The steady-state covariance matrix and the lag-1 steady-state covariance matrix of the SDD model are related via
\begin{align}
    \ccovmat{1}&=\langle\x_{t+\Delta t}\x_t^T\rangle\nonumber\\
    &=[(1-\Delta t_\ctime){\bf I}+\Delta t_\ctime\epsilon\amat]\ccovmat{0}\,.\label{eq:est0}
\end{align}
Defining $\covcov:=\ccovmat{1}\left(\ccovmat{0}\right)^{-1}$, we have for all diagonal elements of $\covcov$
\begin{align}
    \covcovel_{ii}=1-\Delta t_\ctime\,,
\end{align}
when the network has no self-loops. Using the sample covariance matrix and the lag-1 sample covariance matrix as estimators of the steady-state covariance matrix and the lag-1 steady-state covariance matrix, we obtain the estimator
\begin{align}
    \hat\ctime^{-1} = \frac{1-\textrm{med}(\{\hat \covcovel_{ii}\}_{i})}{\Delta t}\,,\nonumber
\end{align}
for the inverse characteristic time $\ctime^{-1}$, where $\textrm{med}(\{\hat \covcovel_{ii}\}_{i})$ denotes the median over all diagonal elements of $\covcov$. 

\section{Pairwise edge measures for the stochastic delay-difference model with order $p>1$}\label{sec:els-proposals2}

In Section \ref{sec:extension}, we derived the process-motif contributions for lag-$k$ covariance in the SDD model of order $p>1$. 
Signal transmission along an edge $e$ with a transmission lag $\maxlag_e$ has a large contribution to the lag-$(1+\maxlag_e)$ covariance from its source node to its target node (see Fig.\,\ref{fig:contributions-cross}). We therefore use a combination of the steady-state covariance matrix with lag $(1+\maxlag_e)$  and a correction term to construct $\hat f(\maxlag_e)$ for inferring edges with transmission lag $\maxlag_e$, i.e., 
\begin{align}
    \hat f(\maxlag_e) = \ccovmat{1+\maxlag_e}-\alpha\ccovmat{\maxlag_e}\,,
\end{align}
for some non-negative $\alpha$.
When one aims to infer edges of any transmission lag (including edges with no transmission lag) up to a maximum transmission lag $\maxlag$, a suitable PEM estimator is 
 the maximum of all $\hat f(\maxlag_e)$ for $\maxlag_e\in\{0,1, \dots, \maxlag\}$, i.e.,
\begin{align}
    f = \max_{\maxlag_e\leq\maxlag} 
    \left[\ccovmat{1+\maxlag_e}_{i,j}-\alpha\ccovmat{\maxlag_e}_{i,j}\right]\,.
\end{align}
Following the same rationale that we used to motivate the \confoundingfactor{} and \reversecausation{} for the order-1 SDD model 
(see Subsection \ref{sec:els-proposals1}), we obtain the PEM estimators 
\begin{align}
\estnamedscore{\confoundingfactor{}}  &:= \max_{\maxlag_e\leq \maxlag}\left[
\sccormat{1+\maxlag_e} \right.\nonumber\\
&\phantom{=\sccormat{1+\maxlag_e}}\left.-\left(1-\Delta t_\ctime+\frac{\mcf_{0,1}(\Delta t_\ctime)}{\mcf_{1,1}(\Delta t_\ctime)}\right)
\sccormat{\maxlag_e}\right]\,,
\end{align}
and
\begin{align}
\estnamedscore{\reversecausation{}}  &:= \max_{\maxlag_e\leq \maxlag}\left[
\sccormat{1+\maxlag_e} - (1-\Delta t_\ctime)\sccormat{\maxlag_e}\right]\,,
\end{align}
for the SDD model of order $\maxlag+1$.

\section{Challenges in inferring large networks}\label{app:shootingstar} 

 \begin{figure}[!tb]
 \centering
 \includegraphics[trim={0.05cm 0cm 0.0cm 3.65cm}, clip,width=0.5\textwidth]{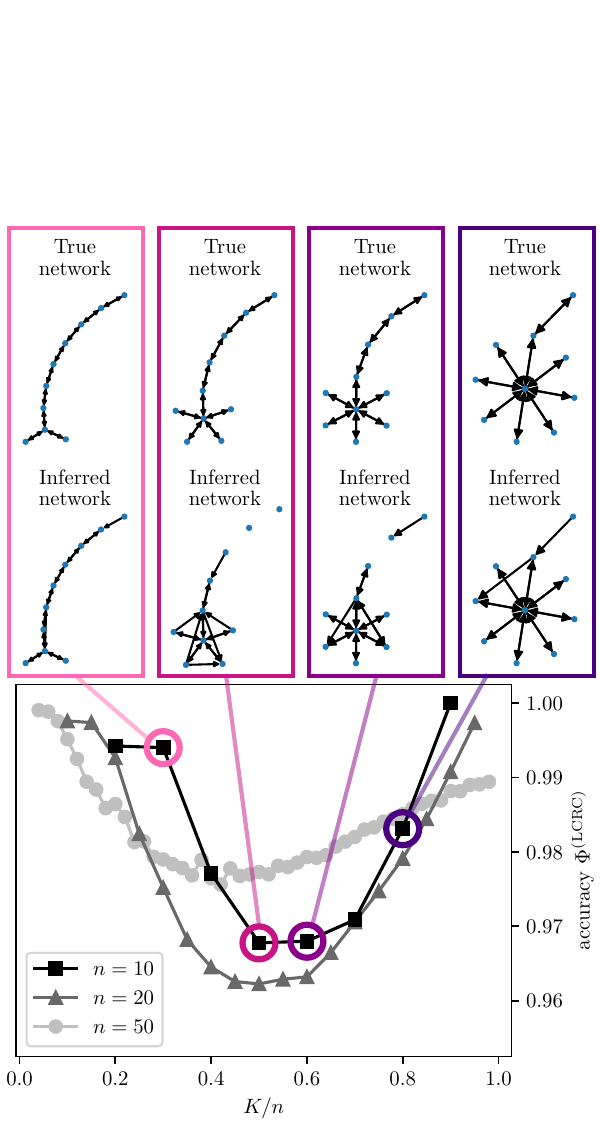}
 \caption{{\bf Inference errors for shooting-star networks.} At the top, we show four exemplary pairs of a ground-truth shooting-star network and a corresponding network inferred from the simulation of the SDD model on the ground-truth network. At the bottom, we show the mean accuracy of network inference via \reversecausation{} from SDD simulations on shooting-star networks. Each data point corresponds to the mean over the accuracy of inferred networks for 100 simulations with inverse sampling
rate $\Delta t = 0.5$, characteristic time $\ctime = 1$, coupling strength
$\epsilon = 0.9$, system noise $\noise = 0.2$, measurement noise $\mnoise = 0$, and $\numsamples = 1000$ observations.}
 \label{fig:appN}
 \end{figure}

To understand some of the challenges that arise in inferring large networks, we characterize structural features that tend to lead to false-postive and/or false-negative edge inferences.

We define a \textit{shooting-star network} to be a graph that one can construct from connecting a path network to a star network by an edge. When each pair of adjacent nodes is connected by a bidirectional edge, one can fully characterize a shooting star network by its number $n$ of nodes and its maximum node degree $K\in\{2,\dots,n-1\}$. In Fig.\,\ref{fig:appN}, we show several examples of small (i.e., $n=10$) shooting-star networks, and corresponding examples of networks inferred via \reversecausation{} from SDD simulations on the shooting-star networks. We also show the mean inference accuracy for shooting-star networks with increasing $K/n$.
A comparison of the inferred networks in Fig.\,\ref{fig:appN} with their corresponding ground-truth networks illustrates that (1) false-positive edge inferences tend to occur for pairs of neighbors of a high-degree node, and (2) false-negative edge inferences tend to occur for pairs of nodes with low degree. Accordingly, the mean inference accuracy for shooting-star networks is high when $K/n$ is close to 0 or close to 1, and is low when $K/n\approx1/2$.

We explain these observations as follows. When the number of edges in a network is known, node pairs are in competition for being assigned an edge in the inferred network because a false-positive edge inference necessitates a false-negative edge inference somewhere else. One can interpret a pair of a false-positive and a false-negative edge inference as an edge in the ground-truth network having lost to a non-edge in the ground-truth network in this competition. Edges between low-degree nodes tend to have the lowest associated PEM values among edges are thus weak competitors in this competition. Conversely, pairs of neighbors of a high-degree node tend to be associated with the highest PEM values among non-edges and thus are strong competitors. A ground-truth network that includes edges between low-degree nodes as well as non-edges between neighbors of a high-degree node (e.g., a shooting-star network with $K/n\approx1/2$) is thus hard to infer with high accuracy, when compared with a ground-truth a network that does not include edges between low-degree nodes (e.g., shooting-star networks with $K/n\approx 1$) or a ground-truth network that does not include non-edges between neighbors of high-degree nodes (e.g., shooting-star networks with $K/n\approx 0$). In conclusion, we observe that the coexistince of (1) edges between low-degree nodes and (2) non-edges between pairs of neighbors of a high-degree node pose a challenge for PEM-based inference methods.

This observation has implications for the accuracy of PEM-based network inference with increasing network size.
A network that includes high-degree nodes with a small local clustering coefficient necessarily also includes non-edges between neighbors of a high-degree node. One can use a node's anticlustering coefficient $\overline{c}_i$ (see Eq.\,\eqref{eq:anticlustering}) as a measure of its tendency to have a high degree and low clustering coefficient. For several graph models, the mean anticlustering coefficient increases with network size (see  Fig.\,\ref{fig:B2}\,(d)). It follows that with increasing network size the number of non-edges that can out-compete edges between low-degree nodes also increases. We expect that this is one of possibly several aspects that make the inference of large networks challenging. 

\bibliography{refs}

\end{document}